\documentclass[letterpaper]{article} %
\usepackage[preprint]{aaai2027}  %
\usepackage[hyphens]{url}  %
\usepackage{graphicx} %
\usepackage{natbib}  %
\usepackage{caption} %
\usepackage{booktabs}
\usepackage{amsmath}
\usepackage{amsfonts}
\usepackage{xspace}
\usepackage{multirow}
\usepackage{makecell}
\usepackage{enumitem}
\usepackage{array}
\usepackage{xurl}

\newcommand{\molbiokg}{MolBioKG\xspace}
\usepackage{amssymb}

\newcommand{\supp}[1]{(See Supp.~Sec.~#1)}
\newcommand{\suppd}[1]{See Supp.~Sec.~#1 for details.}
\title{MolBioKG: Grounding Out-of-Graph Molecules in Biomedical \\ Knowledge Graphs via Multi-Resolution Structural Anchoring}
\author{
    Yiming Zhang\textsuperscript{\rm 1,\rm 2},
    Hikaru Shindo\textsuperscript{\rm 2,*},
    Shuan Chen\textsuperscript{\rm 2},
    Kaushalya Madhawa\textsuperscript{\rm 2},
    Jun Jin Choong\textsuperscript{\rm 2},\\
    Yuna Oikawa\textsuperscript{\rm 1,\rm 2},
    Takashi Fujiwara\textsuperscript{\rm 3,\rm 2},
    Keisuke Ozawa\textsuperscript{\rm 2,*}
}
\affiliations{
    \textsuperscript{\rm 1}The University of Tokyo\\
    \textsuperscript{\rm 2}SB Intuitions\\
    \textsuperscript{\rm 3}Department of AI Systems Medicine, Institute of Science Tokyo\\
    \textsuperscript{*}{\{hikaru.shindo, keisuke.ozawa\}@sbintuitions.co.jp}
}

\begin{document}
\maketitle

\begin{abstract}
Biomedical knowledge graphs (KGs) accelerate drug discovery, but standard pipelines assume query molecules already exist as graph entities, leaving unregistered molecules disconnected. We address this cold-start challenge, termed the \emph{out-of-graph molecule problem}, by introducing \molbiokg. This two-layer system grounds unseen molecules in biomedical evidence via multi-resolution structural anchoring. It connects an index of 2.74 million molecules (represented by scaffolds, fragments, functional groups, and fingerprints) to a 9.6-million-edge KG. Given only a SMILES string, \molbiokg retrieves structurally related graph entities and traverses their biomedical neighborhoods without task-specific training. It features two inference mechanisms: static multi-anchor retrieval using Reciprocal Rank Fusion, and Adapt-KG, a tool-using LLM policy for adaptive traversal. Evaluated across in-graph link recovery, complex multi-hop reasoning, and out-of-graph generalization, \molbiokg outperforms strong baselines. Notably, it raises Hits@10 from 0.585 to 0.876 in multi-hop reasoning and out-of-graph target recall from 0.145 to 0.269, all while ensuring predictions retain traceable structural anchors and source-attributed KG evidence.
\end{abstract}

\section{Introduction}
\label{sec:intro}

Large-scale biomedical knowledge graphs (KGs) are essential for computational drug discovery, integrating diverse evidence across drugs, proteins, and diseases. By mapping biological relationships, these resources enable AI architectures, such as Graph Neural Networks (GNNs), Knowledge-Graph Embeddings (KGEs), and Large Language Models (LLMs), to achieve core objectives, such as drug repurposing, biomedical question answering, and link prediction. Foundational resources (e.g., Hetionet~\citep{Himmelstein2017}, DRKG~\citep{Ioannidis2020}, PrimeKG~\citep{Chandak2023}, and RTX-KG2~\citep{Wood2022}) make these pipelines highly effective~\citep{Bordes2013,Sun2019,Huang2024,Hu2025,Soman2024}. However, a persistent limitation remains: these systems generally assume the query molecule already exists in the graph. %

This dependency restricts multiple algorithmic paradigms. Transductive KGE models require a fixed training vocabulary, preventing direct scoring of unindexed molecules. Zero-shot reasoning models, such as TxGNN~\citep{Huang2024}, generalize to unseen associations among graph-resident entities, not unregistered structures. Inductive methods (GRAIL~\citep{Teru2020}, ULTRA~\citep{Galkin2024}) broaden generalization but still require inference-time relational context. Multimodal approaches (BioBLP~\citep{daza2023bioblp}) incorporate molecular features but remain confined to fixed KG schemas. Consequently, a valid molecule lacking a graph node and incident edges falls entirely outside standard inference capabilities.

This architectural blind spot is increasingly problematic. Molecular generative models~\citep{Tang2024survey,zhang2026autolead}, AI-enabled screening~\citep{Stokes2020}, and lagging database curation cycles~\citep{zdrazil2024chembl,hunter2025drug} routinely yield structures absent from KG snapshots. Consider \emph{foscarbidopa}, a prodrug approved in October 2024 for Parkinson's disease~\citep{Soileau2022foslevodopa}. While its parent carbidopa has been characterized in KGs for decades~\citep{Wurtman1977carbidopa}, a pre-2024 KG snapshot would completely isolate foscarbidopa, even though both share nearly identical Bemis--Murcko scaffolds~\citep{Bemis1996} and BRICS fragments~\citep{Degen2008}. We formulate this cold-start scenario as the \textbf{out-of-graph molecule problem}:  \emph{how can a system connect a molecule to existing biomedical knowledge when it is not yet a registered graph entity?}

Our approach relies on \emph{structural compositionality}: even globally novel molecules consist of familiar, biologically characterized substructures. An unseen molecule may share a Bemis--Murcko scaffold, BRICS fragments, functional groups~\citep{ertl2017algorithm}, or fingerprint environments~\citep{Rogers2010} with established compounds. Operating at varying resolutions 
to structural shifts, these combined signals provide multiple complementary entry points from a raw SMILES string into an existing evidence base.

Building on this insight, we present \molbiokg, a dual-layer architecture designed to ground out-of-graph molecules via multi-resolution structural anchoring (Figure~\ref{fig:overview}(a)). Instead of modifying underlying graph algorithms, \molbiokg uses a specialized molecular layer to map structural signatures to a comprehensive biomedical graph layer. By bridging chemical and biological spaces through identifier equivalence, the system allows an unseen SMILES string to retrieve well-characterized analogues and seamlessly traverse their graph neighborhoods to infer targets, indications, and effects. We provide two complementary inference mechanisms: a training-free retrieval pipeline for robust evidence aggregation, and Adapt-KG, an LLM-driven policy dynamically orchestrating structural retrieval and graph traversal without task-specific fine-tuning.

We evaluate \molbiokg across three reasoning scenarios: (i) \emph{in-graph link recovery}, masking known drug edges to benchmark retrieval against vocabulary-restricted GNNs; (ii) \emph{complex multi-hop reasoning}, using a custom benchmark (MolBioKG-KGQA) to test natural language bridging of molecular and biological data; and (iii) \emph{out-of-graph generalization}, predicting indications and targets for 199 post-2011 approved drugs. These tracks separate standard graph traversal from true cold-start discovery. %

Our primary contributions are as follows:
\begin{itemize}
  \item We introduce \molbiokg, a novel framework that addresses the out-of-graph molecule problem by bridging a large-scale, multi-resolution structural index with a comprehensive biomedical KG.
  \item We develop two complementary reasoning pathways: a static multi-anchor retrieval system using Reciprocal Rank Fusion (RRF)~\citep{RRF}, and Adapt-KG, an LLM-driven policy adaptively orchestrating structural retrieval and KG traversal tools.
  \item We introduce a rigorous three-track evaluation suite encompassing in-graph masked-edge retrieval, cross-layer multi-hop question answering, and true out-of-graph annotation for recently approved drugs.
  \item We empirically demonstrate the complementary strengths of our inference mechanisms. Adapt-KG improves multi-hop Hits@10 from 0.585 to 0.876 over the strongest non-adaptive baseline, while RRF raises out-of-graph LLM-judge recall over zero-shot from 0.186 to 0.239 for indications and 0.145 to 0.269 for targets.  
\end{itemize}

\section{Related Work}
\label{sec:related}

\paragraph{Biomedical KGs for drug discovery.}
Large biomedical KGs, including Hetionet~\citep{Himmelstein2017}, DRKG~\citep{Ioannidis2020}, PrimeKG~\citep{Chandak2023}, and RTX-KG2~\citep{Wood2022}, support KGE methods such as TransE~\citep{Bordes2013}, DistMult~\citep{Yang2015}, and RotatE~\citep{Sun2019}, as well as GNN systems such as TxGNN~\citep{Huang2024} and BioPathNet~\citep{Hu2025}. LLM-based systems further use graph or provenance-structured retrieval to ground reasoning over domain knowledge~\citep{Soman2024, zhang2026provmind}. These approaches are effective when the query entities are available within the graph's entity universe, but standard formulations do not directly address a molecule supplied only as a previously unseen chemical structure.

\paragraph{Generalization beyond training entities.}
Inductive KGE methods infer over entities not observed during training by exploiting their test-time relational neighborhoods. GRAIL~\citep{Teru2020} reasons over enclosing subgraphs, while ULTRA~\citep{Galkin2024} transfers relational patterns across KGs. Multimodal approaches such as BioBLP~\citep{daza2023bioblp} encode molecular structure alongside graph context, and GenSPARC~\citep{zhang2025generalizable} targets transfer across KG schemas. These methods relax the fixed-training-vocabulary assumption, but they generally presuppose that the query entity is represented in an inference graph, accompanied by relational context or a registered entity record. \molbiokg instead begins with a standalone SMILES and constructs an evidence path through structurally related graph entities.

\paragraph{Generative molecules and post-generation annotation.}
Generative models based on variational, autoregressive, and diffusion architectures can propose molecules outside current databases~\citep{Tang2024survey,zhang2026autolead}. Their outputs must subsequently be connected to putative targets, indications, and safety evidence. Existing similarity search can identify close analogues, but it does not by itself organize heterogeneous biomedical relations or support multi-hop reasoning. \molbiokg combines multi-resolution chemical retrieval with an explicit biomedical KG, positioning structural analogues as auditable anchors.

\section{MolBioKG}

\label{sec:method}

\begin{figure*}[t]
\centering
\includegraphics[width=\textwidth]{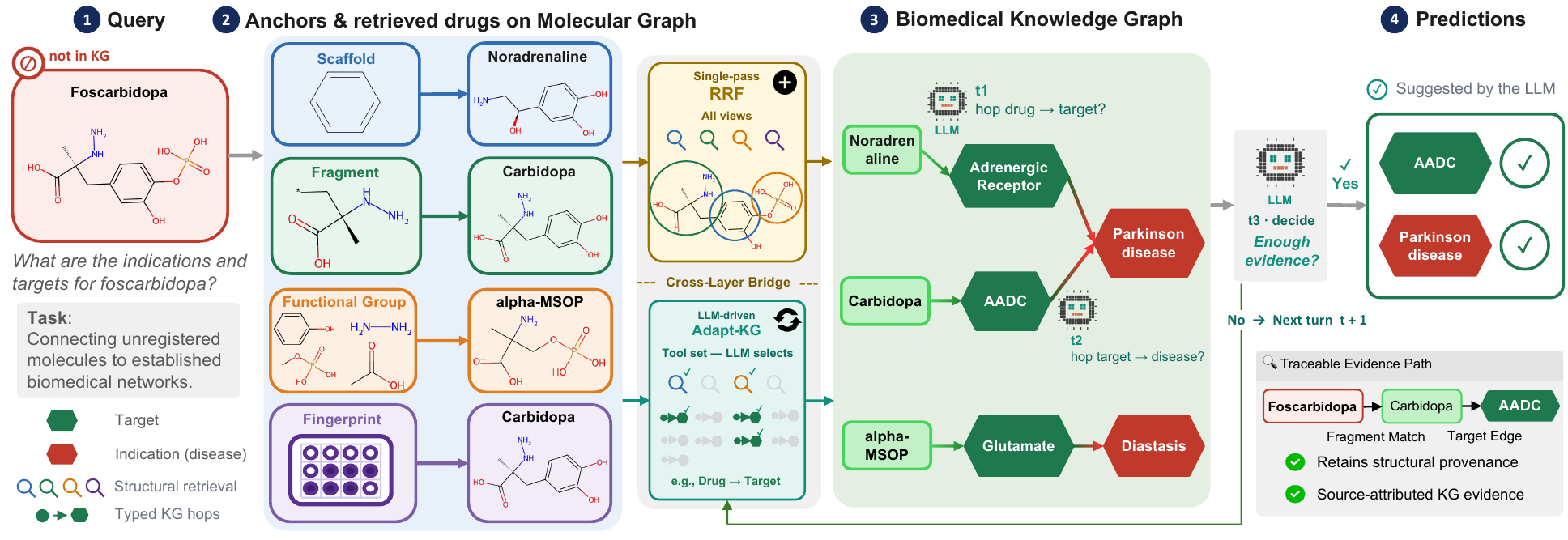}
\caption{\textbf{\molbiokg\ inference pipeline.} \textbf{(1)} An out-of-graph query molecule is given. \textbf{(2)} Four structural anchors retrieve drug analogues, combined via two complementary mechanisms: static RRF fusion over all views, and Adapt-KG, an LLM-driven cross-layer bridge that selects typed KG tools. \textbf{(3)} The KG is traversed from these anchors, with Adapt-KG iterating over multiple turns until the LLM deems evidence sufficient. \textbf{(4)} Evidence across anchors is aggregated, where multi-anchor consensus strengthens a prediction, ensuring every final candidate is backed by a fully traceable structural and graph path.}
\label{fig:overview}
\end{figure*}

Biomedical knowledge graphs (KGs) are limited to analyzing molecules already registered in their network, leaving newly discovered molecules disconnected. To overcome this cold-start problem, we leverage the chemical similarity principle. By identifying known, graph-resident drugs that share structural features with an unseen query molecule, we can transfer their established biological relationships to evaluate the new compound.

Given a standalone molecular representation (e.g., a SMILES string $q$), \molbiokg finds structurally related molecules within the biomedical KG. As shown in Figure~\ref{fig:overview}, a bridging index connects the molecular and biomedical layers. The retrieved molecules act as \emph{structural anchors}, providing entry points into the graph. \molbiokg traverses from these anchors to gather and rank candidate evidence, such as targets and indications, for the new molecule.
Crucially, \molbiokg ensures traceable hypothesis generation by treating the anchors' biological relationships as ranked candidate evidence rather than definitive clinical proof.

\subsection{MolBioKG Architecture}
\label{sec:system}

To connect an unregistered molecule to existing biological knowledge, \molbiokg uses a decoupled, two-layer architecture. First, a \emph{molecular layer} analyzes the new query's chemical structure to retrieve similar, known molecules. Second, a \emph{biomedical layer} provides the established biological relationships (such as targets and indications) for those known entities. A \emph{bridge index} maps the chemical identifiers to these biomedical nodes. This design allows the system to infer biological evidence for a completely new molecule without needing it to be present in the underlying knowledge graph.

\paragraph{Molecular structure layer.} This structural index locates out-of-graph queries using 2.74 million molecules mapped from RTX-KG2c. These structures originate from canonical registries, such as PubChem~\citep{kim2025pubchem}, ChEMBL~\citep{zdrazil2024chembl}, and ChEBI~\citep{degtyarenko2007chebi}. For robust retrieval, the layer indexes each molecule at multiple resolutions: Bemis--Murcko scaffolds~\citep{Bemis1996} for the central framework, BRICS fragments~\citep{Degen2008} for substructures, functional groups~\citep{ertl2017algorithm} for broad characteristics, and ECFP4 fingerprints~\citep{Rogers2010} for local environments. These complementary views capture distinct aspects of chemical similarity, recovering pharmacologically relevant anchors that a single representation might miss.

\paragraph{Biomedical evidence layer.} This layer serves as the biological ground truth for the system, storing established clinical and molecular relationships. To ensure that candidate predictions are backed by empirical evidence rather than chemical similarity alone, we populate this layer using RTX-KG2c~\citep{Wood2022}, a comprehensive biomedical knowledge graph that integrates over 70 public biological databases into a single standardized network. It is structured as a network of 12 drug-discovery-relevant node categories (e.g., drugs, diseases, genes, proteins) connected by specific predicates, such as drug--target and protein-interaction evidence. The resulting graph contains approximately 860,000 nodes and 9.6 million edges.

\paragraph{Cross-layer bridge.} This component acts as the translation mechanism between the chemical and biological domains. Because the two distinct layers rely on different identifier systems, a retrieved structural analogue cannot natively query the biomedical graph. To resolve this, the bridge maps molecular-layer results to their corresponding biomedical entities using shared identifiers, universal chemical barcodes (matching InChIKeys), and explicit equivalence edges within the graph (RTX-KG2 \texttt{same\_as} links). Isolating this translation step ensures that the system safely grounds structural anchors without introducing new biological claims, keeping the overall architecture  modular. \suppd{1} %

\subsection{Static Multi-Anchor Retrieval}
\label{sec:mol_layer}

To establish reliable entry points into the biomedical layer, this inference mechanism employs a static, multi-faceted retrieval strategy. Because chemical similarity spans multiple scales, relying on a single structural representation often yields matches that are either too narrow to be useful or too broad to be accurate. To overcome this, \molbiokg processes an unseen query SMILES $q$ by independently searching across four distinct structural views and aggregating the resulting anchors, ensuring a robust connection to the graph.

\paragraph{Structural anchor retrieval.} To execute this search, the retrieval module evaluates graph-resident molecules against the query using four complementary dimensions of chemical similarity. Specifically, the \textbf{Scaffold} view favors molecules with the same core framework and falls back to broader ring-based matches. The \textbf{Fragment} view identifies molecules that share BRICS substructures while reducing the influence of very common fragments. The \textbf{Functional group (FG)} view compares the types and frequencies of chemical groups, and the \textbf{Fingerprint (FP)} view uses ECFP4 Tanimoto similarity to capture overall agreement across local atomic environments. Ties and residual ordering within a channel are further resolved using each candidate's connectivity in the biomedical graph \supp{5}. 

\paragraph{Rank fusion and evidence aggregation.}
Because the four views produce scores with different meanings, we combine their ranked lists rather than comparing their raw scores. We use Reciprocal Rank Fusion (RRF)~\citep{RRF}, which gives more support to molecules that appear near the top of one or more lists: $\mathrm{RRF}(a)=\sum_{r\in\mathcal{R}}\frac{1}{c+\operatorname{rank}_{r}(a)}$,
where $\mathcal{R}$ is the set of available rankings and $c$ is a fixed offset. We exclude the query itself using canonical-SMILES and InChIKey checks. The highest-ranked anchors are then mapped through the bridge index, and their biomedical neighbors are collected as candidate relations. Each candidate retains the supporting anchor, structural score, KG predicate, and source record. This final aggregation step turns chemical similarity into traceable biomedical evidence while preserving the distinction between an analogue and the query molecule.

\subsection{Adapt-KG: LLM-Guided Graph Traversal}
\label{sec:adaptkg}

To answer complex biomedical queries requiring multi-hop reasoning, a system must dynamically compose evidence. Since static RRF's fixed retrieval cannot navigate multi-hop reasoning chains, we introduce Adapt-KG. This agentic framework uses a pretrained Large Language Model (LLM) to dynamically orchestrate structural-retrieval and KG-traversal actions through sequential decision-making, requiring no parameter updates.

At each turn, the LLM selects actions, observes evidence, and continues or answers. It uses four structural tools for out-of-graph queries to find entry points, and nine KG-hop tools for natural-language queries to compose evidence. Retaining exact graph identifiers ensures the final response explicitly cites its reasoning path \supp{6}.

\paragraph{Task-specific inference.} %
For \emph{in-graph link recovery}, static RRF directly ranks the biomedical entities collected from structural anchors, with no LLM in the loop. In \emph{out-of-graph generalization}, evidence for a molecule with no biomedical annotations comes from either static RRF or Adapt-KG's own retrieval loop; in both cases, an LLM synthesizes candidate targets, indications, or other properties from the resulting context. Finally, \emph{complex multi-hop reasoning} uses typed traversal tools to connect relations across the molecular and biomedical layers, with Adapt-KG selecting and interleaving these tools for at most six turns, with KGQA policies, KGQA benchmark construction, and out-of-graph protocols detailed in Supp. Sec. 7, 9, and 11, respectively.

\section{Experiments}
\label{sec:experiments}

To evaluate \molbiokg, we design three experiments: \emph{In-graph link recovery} compares retrieval against standard graph models using known drugs; \emph{Complex multi-hop reasoning} tests cross-layer evidence composition; and \emph{out-of-graph generalization} evaluates the core cold-start challenge where queries lack biomedical annotations.

\subsection{In-Graph Link Recovery}
\label{sec:retrieval_benchmark}

This experiment evaluates how well static structural retrieval can predict missing edges between known entities. The task is formulated as a ranking problem: given a test drug, the system must score and rank a vast pool of candidate nodes (e.g., all possible diseases or targets) to recover the drug's true, held-out relations. To ensure a fair comparison with standard graph models, \molbiokg\ is restricted to using structural anchors from training drugs whose relations remain visible in the graph. Unlike the cold-start experiment, the query molecule is a known graph entity, allowing us to isolate retrieval performance under controlled conditions.

\paragraph{Tasks and generalization regimes.}
We define four specific link prediction tasks spanning key drug-discovery decisions: predicting whether a drug could treat (T1) or is contraindicated in (T2) a disease (DrugCentral~\citep{avram2023drugcentral}), which protein the drug interacts with (T3; DrugBank~\citep{knox2024drugbank}, ChEMBL~\citep{zdrazil2024chembl}, and DrugCentral), and whether the drug directionally increases or decreases a given pharmacological effect (T4; 58,261 object--direction labels from RTX-KG2c).
We evaluate these tasks across four data splits to measure robustness to distribution shift: L1 is a standard random split; L2 withholds complete Bemis--Murcko scaffold groups from the training set; L3 withholds Butina clusters~\citep{butina1999unsupervised}; and L4 withholds entire MONDO~\citep{vasilevsky2025mondo} disease areas for T1 and T2. Crucially, L1--L3 all perturb the \emph{chemical} novelty of the query drug (random, scaffold, and fingerprint-cluster holdouts on the drug side), whereas L4 holds the query chemistry in-distribution and instead withholds entire disease areas on the \emph{answer} side. 
L4 therefore probes an orthogonal generalization axis: whether structural anchoring recovers indications and contraindications for disease families unseen during training, rather than for chemically novel query molecules \supp{2}. 

\paragraph{Baseline and metrics.}
We compare against TxGNN~\citep{Huang2024}, a state-of-the-art graph foundation model for drug repurposing. Its zero-shot generalization is over diseases within a fixed drug vocabulary, and as a transductive model it cannot score drugs outside that vocabulary. We therefore evaluate both methods on the intersection of test drugs with TxGNN's vocabulary, ensuring a fair comparison on the same molecules. We report Hits@1, Hits@10, and mean reciprocal rank (MRR).

\begin{table*}[t]
\centering
\caption{\textbf{Static structural anchoring is highly competitive for missing link prediction.} To establish this baseline, \molbiokg\ is evaluated against TxGNN, a state-of-the-art graph foundation model, on a shared entity vocabulary across four relation types (T1: indication, T2: contraindication, T3: target, T4: directional effect). Generalization is tested under random (L1), scaffold (L2), ECFP4-cluster (L3), and disease-area (L4) distribution shifts. \textbf{Bold} indicates the highest score per column.}
\label{tab:results_main}
\footnotesize
\setlength{\tabcolsep}{3.5pt}
\begin{tabular}{ll rrr rrr rrr rrr}
\toprule
& & \multicolumn{3}{c}{L1} & \multicolumn{3}{c}{L2} & \multicolumn{3}{c}{L3} & \multicolumn{3}{c}{L4} \\
\cmidrule(lr){3-5}\cmidrule(lr){6-8}\cmidrule(lr){9-11}\cmidrule(lr){12-14}
Task & Method & H@1 & H@10 & MRR & H@1 & H@10 & MRR & H@1 & H@10 & MRR & H@1 & H@10 & MRR \\
\midrule
\multirow{2}{*}{T1}
 & TxGNN~\citep{Huang2024}  & .542 & .843 & .656 & \textbf{.602} & \textbf{.847} & \textbf{.695} & \textbf{.618} & .841 & \textbf{.700} & .570 & .859 & .666 \\
 & \molbiokg (ours)        & \textbf{.660} & \textbf{.885} & \textbf{.750} & .387 & .794 & .543 & .563 & \textbf{.879} & .679 & \textbf{.582} & \textbf{.875} & \textbf{.691} \\
\midrule
\multirow{2}{*}{T2}
 & TxGNN~\citep{Huang2024}  & \textbf{.604} & .727 & \textbf{.670} & \textbf{.690} & \textbf{.795} & \textbf{.746} & \textbf{.592} & \textbf{.715} & \textbf{.659} & \textbf{.717} & .861 & \textbf{.790} \\
 & \molbiokg (ours)        & .416 & \textbf{.812} & .536 & .246 & .778 & .395 & .307 & .709 & .430 & .472 & \textbf{.900} & .602 \\
\midrule
\multirow{2}{*}{T3}
 & TxGNN~\citep{Huang2024}  & .235 & .757 & .402 & .295 & \textbf{.780} & .460 & .293 & \textbf{.778} & .452 & --- & --- & --- \\
 & \molbiokg (ours)        & \textbf{.710} & \textbf{.855} & \textbf{.763} & \textbf{.320} & .718 & \textbf{.465} & \textbf{.548} & .773 & \textbf{.629} & --- & --- & --- \\
\midrule
\multirow{2}{*}{T4}
 & TxGNN~\citep{Huang2024}  & .285 & .748 & .447 & .302 & .780 & .468 & .335 & .753 & .486 & --- & --- & --- \\
 & \molbiokg (ours)        & \textbf{.739} & \textbf{.898} & \textbf{.802} & \textbf{.346} & \textbf{.865} & \textbf{.531} & \textbf{.607} & \textbf{.892} & \textbf{.708} & --- & --- & --- \\
\bottomrule
\end{tabular}
\end{table*}

\subsection{Complex Multi-Hop Reasoning}
\label{sec:llm_setup}

This experiment evaluates the system's ability to answer complex, natural-language biomedical questions that span multiple domains. Unlike direct link prediction, these queries require multi-hop reasoning—the ability to dynamically chain together evidence across several different entity types (e.g., inferring shared molecular functional groups among drugs that treat a specific disease). We formulate this as a Question Answering (QA) task to test whether an LLM can navigate both the chemical and biological layers of the graph, comparing our adaptive traversal strategy (Adapt-KG) against baselines that rely on fixed retrieval plans.

\paragraph{Benchmark.}
To evaluate the multi-hop reasoning capability, we construct \textbf{MolBioKG-KGQA}, a dataset of 460 complex questions spanning 2--5 reasoning hops. The questions are categorized into five path families that cross the molecular-biological boundary: Disease-to-Target, Target-to-Disease, Disease-to-Functional-Group, Target-Modulation-to-Functional-Group, and Functional-Group-to-Phenotype. To ensure linguistic variety, the phrasing for each question is sampled from five human-written paraphrases. Because all questions and answers are rigorously instantiated from the frozen KG (with each query guaranteeing at least three gold endpoints and five intermediate drugs),
this benchmark explicitly measures a model's ability to retrieve and compose graph evidence without relying on memorized clinical knowledge. \suppd{9}

\paragraph{Baselines and metrics.}
All retrieval-based methods use the same nine typed traversal tools, so the comparison isolates how evidence is selected and organized. Zero-shot uses no tools. RAG plans a small set of transitions and answers from flat retrieved context, whereas Graph-RAG~\citep{Edge2024} preserves typed and aggregate structure. Graph-CoT~\citep{jin2024graph} chooses one transition at a time, and Think-on-Graph~\citep{Sun2024} maintains a beam of candidate entities. Adapt-KG adaptively selects among the same tools. We report Hits@1/5/10, and entity-set macro-F1. Mol-only and Bio-only variants restrict access to one graph layer, testing whether successful reasoning depends on their connection \supp{7}. %
RRF and TxGNN are not included: RRF's structural-similarity functions have no counterpart for the non-molecular entities (disease, gene) this benchmark starts from, and TxGNN has no natural-language interface and scores only fixed drug-disease pairs; its multi-hop paths are a post-hoc explanation via GraphMask edge-importance scoring of a precomputed prediction, not a mechanism to compose evidence for a new query.

\subsection{Out-of-Graph Generalization}
\label{sec:oog_setup}

The final experiment turns to the setting for which \molbiokg is primarily designed: annotating a molecule when its own graph record provides no biomedical relations.

\paragraph{Benchmark.}
The out-of-graph problem is best tested prospectively, on molecules the system could not have seen, rather than on a hand-curated holdout. We therefore impose a temporal cutoff: both the molecular and biomedical layers are restricted to ChEMBL release~10 (June 2011), and evaluation drugs are drawn from compounds first approved afterward. This construction needs no manual leakage check, since a drug approved after the cutoff cannot appear in the pre-cutoff KG by definition. It yields 199 approved drugs (2012--2024), all confirmed absent from the 2011 snapshot at both the structural and biomedical level, with reference indications and targets (4.0 and 1.3 per drug on average) drawn directly from ChEMBL mechanism-of-action records, independent of the KG. \suppd{11} %

\paragraph{Baselines and metrics.}
Zero-shot measures prediction from the SMILES alone, while Random Context controls for the effect of providing an equally formatted but structurally irrelevant candidate list. The four single-anchor strategies test each structural view independently. RRF combines all four rankings, and Adapt-KG chooses which views to consult. We additionally include Label Propagation, which removes the LLM synthesis step and predicts directly from RRF-weighted anchor votes. 

We report exact recall (Rec\textsubscript{ex}) using normalized surface-form matching and LLM-judge recall (Rec\textsubscript{llm}) for semantically equivalent predictions. The former is deterministic but sensitive to naming differences; the latter is more tolerant to paraphrases but depends on the judge model. 

\subsection{Results}
\label{sec:results}

We report results in the same order as the three evaluation questions above. The first experiment establishes the behavior of static structural retrieval under controlled graph coverage, the second evaluates adaptive composition over multiple hops, and the third tests genuine out-of-graph annotation.

\subsubsection{In-Graph Link Recovery}
\label{sec:results_retrieval}

\begin{figure}[t]
\centering
\includegraphics[width=\columnwidth]{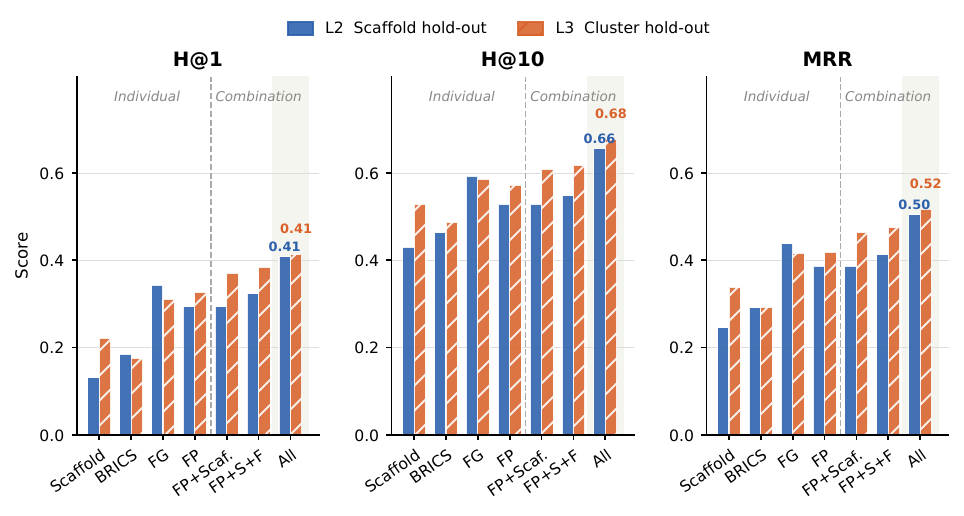}
\caption{\textbf{Combining multiple structural views provides complementary evidence that outperforms any single anchor.} Anchor ablation for indication recovery (T1) under scaffold (L2) and ECFP4-cluster (L3) holdouts ($K=5$). While Functional Group and Fingerprint perform strongly in individual regimes, fusing all views via RRF yields the highest overall performance across different distribution shifts.}
\label{fig:ablation}
\end{figure}

\begin{table}[t]
\centering
\caption{\textbf{Adapt-KG outperforms existing graph-reasoning baselines on natural-language multi-hop questions.} Performance on the MolBioKG-KGQA benchmark, where all tool-using methods share the same nine typed traversal tools and differ only in how they select transitions. The Mol-only and Bio-only variants act as layer-restriction ablations, showing that successful reasoning requires traversing both the molecular and biomedical domains (Best scores in bold). %
}
\label{tab:llm}
\small
\setlength{\tabcolsep}{4pt}
\resizebox{\columnwidth}{!}{%
\begin{tabular}{l rrrr @{\hspace{6pt}} r @{\hspace{4pt}} r}
\toprule
& \multicolumn{4}{c}{Full KG}
& \multicolumn{1}{c}{Mol-only}
& \multicolumn{1}{c}{Bio-only} \\
\cmidrule(lr){2-5}\cmidrule(lr){6-6}\cmidrule(lr){7-7}
Method & H@1 & H@5 & H@10 & F1 & H@10 & H@10 \\
\midrule
Zero-shot       & .059 & .163 & .196 & .029 & ---  & ---  \\
RAG       & .328 & .385 & .407 & .259 & .204 & .396 \\
Think-on-Graph & .337 & .470 & .476 & .165 & \textbf{.248} & .389 \\
Graph-CoT      & .435 & .570 & .585 & .332 & .133 & .298 \\
Graph-RAG      & .454 & .557 & .583 & .298 & .000 & .394 \\
Adapt-KG       & \textbf{.776} & \textbf{.872} & \textbf{.876} & \textbf{.570} & .180 & \textbf{.628} \\
\bottomrule
\end{tabular}%
}
\end{table}
\paragraph{Comparison with TxGNN.}
\molbiokg provides strong broad-recall performance on graph-resident drugs, exceeding TxGNN in Hits@10 on 9 of the 14 comparable task--regime combinations (Table~\ref{tab:results_main}). The gains are most consistent for directional effects (T4), where \molbiokg leads under every available regime, and for indications (T1), where it leads under L1, L3, and L4. Contraindication recovery (T2) is more mixed: TxGNN achieves higher Hits@1 and MRR across all regimes, whereas \molbiokg yields higher Hits@10 under L1 and L4. Thus, while TxGNN ranks the correct contraindication first more reliably, \molbiokg recovers a broader set of relevant candidates within the top ten.

\paragraph{Generalization under structural shift.}
Across regimes, comparative performance is strongly shaped by structural distribution shift.
In particular, removing exact scaffold overlap (L2) represents the most challenging setting for \molbiokg: TxGNN leads on T1, T2, and T3 because structural retrieval must rely on less specific fragment, functional-group, and fingerprint evidence. However, performance recovers under ECFP4-cluster holdout (L3) for T1 and T4, where clustering by fingerprint similarity, unlike scaffold holdout, does not prevent a test drug from sharing its exact scaffold with training data (Detailed results in Supp. Sec. 3).

\paragraph{Fusing structural views is complementary.}
These regime-dependent variations highlight why combining multiple structural representations is essential. Our ablation analysis confirms that the four structural views contribute complementary evidence across distinct shifts (Figure~\ref{fig:ablation}). Functional Group serves as the strongest individual view under scaffold shift (L2), whereas Fingerprint performs best under cluster shift (L3), demonstrating that no single feature family dominates both regimes. Fusing all four views yields the highest overall performance throughout, reaching Hits@1 of 0.408 under L2 and 0.414 under L3. The benefit of RRF therefore stems from integrating distinct structural relationships rather than repeatedly measuring a single similarity signal.

\subsubsection{Complex Multi-Hop Reasoning}
\label{sec:results_llm}

\paragraph{Adaptive traversal drives reasoning gains.}
When evaluating complex multi-hop queries, adaptive graph traversal proves essential for effective evidence synthesis. Adapt-KG achieves the top performance on MolBioKG-KGQA, reaching 0.876 Hits@10 and 0.570 macro-F1 (Table~\ref{tab:llm}). By comparison, the strongest non-adaptive policy, Graph-CoT, reaches only 0.585 Hits@10 and 0.332 macro-F1, while even flat RAG, the least structured of the tool-using policies, already exceeds Zero-shot across every full-KG metric, roughly doubling its Hits@10 (0.407 vs.\ 0.196). Merely retrieving larger volumes of entities is therefore insufficient; the primary advantage stems from dynamically selecting transitions aligned with the question's specific entity types and preserving the resulting evidence path.

\paragraph{Chemical and biological layers act synergistically.}
Beyond traversal strategy, successful multi-hop reasoning depends heavily on the synergy between chemical and biological information. Restricting access to either individual graph layer significantly reduces performance relative to the integrated full graph, for every method. Under Mol-only restriction, Think-on-Graph is the most robust policy (0.248 Hits@10), ahead of Adapt-KG (0.180); under Bio-only restriction, Adapt-KG remains the strongest policy (0.628 Hits@10). Both single-layer outcomes for Adapt-KG fall well short of its 0.876 full-KG baseline. These comparisons highlight the critical value of cross-layer connectivity, though they should be interpreted as layer-restriction tests since withholding a layer also removes its evidence. Additional backbones are compared in Supp. Sec. 13. %
  
\subsubsection{Out-of-Graph Generalization}

\label{sec:results_oog}

\begin{table}[t]
\centering
\caption{\textbf{Fusing structural anchors yields the highest out-of-graph annotation recall.} Performance is evaluated using gpt-oss-120b, reporting mean$\pm$std over 3 runs. The combined RRF strategy consistently outperforms single-anchor views and baseline contexts 
(Best scores in bold).
}
\label{tab:oog_grounding}
\small
\resizebox{\columnwidth}{!}{%
\begin{tabular}{l rr rr}
\toprule
& \multicolumn{2}{c}{Exact Recall (Rec\textsubscript{ex})}
& \multicolumn{2}{c}{LLM-Judge Recall (Rec\textsubscript{llm})} \\
\cmidrule(lr){2-3}\cmidrule(lr){4-5}
Strategy & Ind. & Tgt. & Ind. & Tgt. \\
\midrule
Zero-shot (no KG)
  & 0.085$\pm$0.003 & 0.144$\pm$0.016 & 0.186$\pm$0.006 & 0.145$\pm$0.016 \\
Random Context
  & 0.086$\pm$0.006 & 0.114$\pm$0.017 & 0.181$\pm$0.009 & 0.115$\pm$0.016 \\
Label Propagation
  & 0.038$\pm$0.000 & 0.156$\pm$0.000 & 0.105$\pm$0.000 & 0.161$\pm$0.000 \\
\midrule
Fingerprint (FP)
  & 0.105$\pm$0.003 & 0.206$\pm$0.008 & 0.221$\pm$0.013 & 0.208$\pm$0.009 \\
Scaffold
  & 0.115$\pm$0.008 & 0.184$\pm$0.008 & 0.208$\pm$0.009 & 0.186$\pm$0.009 \\
BRICS Fragment
  & 0.116$\pm$0.007 & 0.215$\pm$0.010 & 0.215$\pm$0.006 & 0.219$\pm$0.010 \\
Funct.\ Group (FG)
  & 0.119$\pm$0.002 & 0.211$\pm$0.013 & 0.223$\pm$0.003 & 0.212$\pm$0.012 \\
RRF
  & \textbf{0.125$\pm$0.003} & \textbf{0.266$\pm$0.008} & \textbf{0.239$\pm$0.005} & \textbf{0.269$\pm$0.009} \\
Adapt-KG
  & 0.123$\pm$0.006 & 0.216$\pm$0.014 & 0.232$\pm$0.009 & 0.218$\pm$0.015 \\
\bottomrule
\end{tabular}%
}
\end{table}

\begin{table}[t]
\centering
\caption{\textbf{Retrieved structural anchors enable novel biomedical synthesis beyond direct copying.} Attribution and Novel Recall for out-of-graph annotation. Novel Recall evaluates recovery of gold entities absent from retrieved context. $n$ represents the mean number of target entities.}
\label{tab:oog_attribution}
\resizebox{\columnwidth}{!}{
\begin{tabular}{l r rr}
\toprule
Strategy & KG-Attr. & Novel Ind.\ Rec.\ ($n$) & Novel Tgt.\ Rec.\ ($n$) \\
\midrule
Zero-shot (no KG) & 0.000 & 0.085 (199) & 0.144 (199) \\
Random Context   & 0.045 & 0.067 (199) & 0.106 (194) \\
\midrule
Fingerprint (FP)   & 0.182 & 0.088 (197) & 0.152 (183) \\
Scaffold           & 0.023 & 0.111 (199) & 0.163 (194) \\
BRICS Fragment     & 0.087 & 0.108 (199) & 0.168 (188) \\
Funct.\ Group (FG) & 0.052 & 0.113 (199) & 0.193 (194) \\
RRF         & 0.200 & 0.106 (197) & 0.199 (181) \\
Adapt-KG           & 0.118 & 0.113 (197) & 0.174 (182) \\
\bottomrule
\end{tabular}
}
\end{table}

\paragraph{Structural relevance drives out-of-graph recovery.}
While the preceding evaluations demonstrate effective evidence synthesis for established graph entities, out-of-graph generalization directly assesses whether \molbiokg can generate actionable, traceable biomedical annotations for unregistered compounds. In this strict cold-start regime, we interpret the reported recall metrics as evidence-backed candidate recovery against reference indications and targets, measuring structural retrieval quality rather than a definitive confirmation of biological validity.
Providing additional entities is not sufficient unless they are structurally related to the query. Random Context reaches only 0.181 indication and 0.115 target Rec\textsubscript{llm}, both below Zero-shot at 0.186 and 0.145 (Table~\ref{tab:oog_grounding}). RRF gives the strongest recovery on every column, reaching 0.239 indication and 0.269 target Rec\textsubscript{llm} (0.125 and 0.266 Rec\textsubscript{ex}), ahead of every single-anchor view. The single Fingerprint (FP) anchor is functionally analogous to classical ECFP4--Tanimoto similarity search, serving as a classical chemoinformatics reference point. RRF surpasses it on every column (0.269 vs.\ 0.208 target Rec\textsubscript{llm}), showing that fusing structural resolutions recovers annotations single-fingerprint search alone misses. Fusing structural views is therefore the more reliable strategy once evaluation is anchored to a genuine temporal cutoff, where no candidate can share an exact pre-cutoff structure with the query.

\paragraph{LLM synthesis adds value beyond anchor voting.}
While static RRF provides strong structural anchors, directly predicting from them is insufficient. A Label Propagation baseline, which bypasses the LLM and votes directly on the RRF-fused anchors, reaches only 0.038 exact indication recall and 0.156 exact target recall. By contrast, allowing the LLM to synthesize an answer from the exact same retrieved evidence boosts performance to 0.125 and 0.266, respectively. This substantial gap, particularly for indications, highlights that the LLM contributes far more than simple output formatting; it actively reconciles diverse nomenclatures and integrates evidence spanning several retrieved anchors.

\paragraph{Gains reflect genuine synthesis rather than verbatim extraction.}
While the LLM actively integrates retrieved evidence, its success is not driven merely by copying entities in the context window. We demonstrate this using \emph{Novel Recall}, which restricts evaluation to gold entities absent from the retrieved context. Under this metric, RRF maintains a strong advantage over Random Context for both indications (0.106 vs.\ 0.067) and targets (0.199 vs.\ 0.106) (Table~\ref{tab:oog_attribution}). Across the individual strategies, RRF secures the highest Novel Target Recall, whereas Functional Group and Adapt-KG tie for the highest Novel Indication Recall at 0.113. %
This demonstrates that structurally retrieved anchors provide essential context that supports genuine biomedical synthesis, rather than surface-level entity repetition.

\paragraph{Reasoning complexity dictates the optimal inference strategy.}
Comparing the two inference mechanisms reveals that they excel in different contexts. For this focused annotation task, static RRF outperforms Adapt-KG, which reaches 0.232 indication and 0.218 target Rec\textsubscript{llm} despite requiring more inference steps. In multi-hop KGQA, an opposite pattern emerges: Adapt-KG benefits from dynamically choosing typed transitions as the evidence path unfolds. These results indicate that static fusion is the stronger default when a task primarily requires relevant structural evidence, whereas adaptive traversal is essential for compositional questions. See Supp. Sec. 12 for additional backbones.

\paragraph{Case studies validate multi-resolution fusion and semantic evaluation.}
Tracing individual molecules through their retrieved analogues reveals why no single structural view is uniformly reliable. For example, Ensartinib has no scaffold match, yet BRICS identifies substructures shared with Crizotinib and recovers relevant ALK evidence \supp{15}. Conversely, other cases demonstrate fingerprints capturing broader similarity when exact local structures do not match. Alongside these structural variations, disease synonyms and ontology granularity often obscure true performance under rigid metrics (See Supp. Sec. 14).
Overall, these observations validate our dual approach: fusing multiple structural resolutions for robust retrieval, and using semantic recall to accurately measure it.

\section{Conclusion}
\label{sec:conclusion}

Before concluding, we note a few limitations. \molbiokg's performance decreases when query molecules lack exact scaffold overlap with known entities, and its true accuracy can occasionally be obscured by rigid evaluation metrics. Future work could address these constraints by integrating more flexible structural embeddings and nuanced semantic evaluation paradigms.

To conclude, we introduced \molbiokg to resolve the out-of-graph molecule problem by grounding unregistered compounds within biomedical knowledge graphs. By integrating multi-resolution structural anchors with source-attributed graph evidence, it bridges novel chemical structures and established biological knowledge without retraining models.
Our evaluations demonstrate the complementary strengths of \molbiokg's dual inference mechanisms. Static rank fusion consistently outperforms state-of-the-art graph models in link recovery, while LLM-guided adaptive traversal nearly doubles retrieval accuracy for complex multi-hop reasoning. Crucially, on a strict temporal hold-out of recently approved drugs, the system substantially elevates out-of-graph recall for both indications and targets.
Overall, \molbiokg transforms isolated molecular structures into traceable candidate evidence, establishing a transparent and scalable foundation for AI-driven drug discovery.

\bibliography{aaai2027}

\clearpage

\setcounter{secnumdepth}{2}
\setcounter{section}{0}
\renewcommand{\thetable}{S\arabic{table}}
\renewcommand{\thefigure}{S\arabic{figure}}
\setcounter{table}{0}
\setcounter{figure}{0}

\begin{center}
{\LARGE\sc MolBioKG: Supplementary Material}
\end{center}
\vspace{0.15in}

This supplement is organized around three goals. Sections~\ref{app:construction}, \ref{app:retrieval}, and \ref{app:adaptkg_prompts} describe how \molbiokg is constructed and how its static multi-anchor retrieval and Adapt-KG policy are implemented. Sections~\ref{app:tasks}, \ref{app:baselines}, \ref{app:kgqa_construction}, and \ref{app:oog_construction} define the three evaluation settings, their protocols, and comparison baselines. The remaining sections provide complete results and ablations, inference-cost analyses, robustness experiments, normalization studies, and qualitative examples that complement the main paper.

\section{MolBioKG Construction}
\label{app:construction}

This section explains how the two-layer knowledge system is constructed. The biomedical layer provides source-attributed biological and clinical relations, the molecular layer makes chemical structures searchable at multiple resolutions, and the bridge index connects equivalent molecule records across the two layers.

\subsection{Biomedical Layer}

The biomedical layer supplies the relations used to support candidate answers. We derive it from RTX-KG2c v2.10.0~\citep{Wood2022}, retaining the 12 drug-discovery-relevant node categories and the eight relation types summarized in Table~\ref{tab:app_bio_predicates}. We call a node \emph{bio-annotated} when it participates in at least one of these relations. This definition determines whether a molecular record can provide biomedical evidence after cross-layer mapping.

\begin{table*}[t]
\centering
\caption{Biomedical relation types retained for retrieval, bridging, and evaluation.}
\label{tab:app_bio_predicates}
\small
\begin{tabular}{p{0.62\textwidth}p{0.28\textwidth}}
\toprule
Biolink relation & Role in \molbiokg \\
\midrule
\texttt{treats\_or\_applied\_or\_studied\_to\_treat} & Treatment and indication evidence \\
\texttt{physically\_interacts\_with}; \texttt{directly\_physically\_interacts\_with} & Drug--target interaction evidence \\
\texttt{contraindicated\_in} & Contraindication evidence \\
\texttt{causes} & Causal and adverse-effect evidence \\
\texttt{affects} & Directional pharmacological effects \\
\texttt{gene\_associated\_with\_condition} & Gene--disease associations \\
\texttt{biomarker\_for} & Biomarker relations \\
\bottomrule
\end{tabular}
\end{table*}

\subsection{Molecular Layer}

The molecular layer provides several ways to locate analogues when the query itself has no biomedical record. It contains 2.74 million molecules whose SMILES originate primarily from PubChem, ChEMBL, and ChEBI records in RTX-KG2c. We represent each molecule through four complementary structural views:

\begin{itemize}
  \item \textbf{Scaffold.} RDKit computes each molecule's Bemis--Murcko scaffold, and molecules with the same canonical scaffold SMILES share a scaffold node. A coarser \emph{generic scaffold} is obtained by replacing atom and bond types with generic carbon and single-bond representations and removing aromaticity. This representation provides a fallback when no other molecule shares the exact decorated scaffold.
  \item \textbf{BRICS fragment.} Each molecule is decomposed into BRICS fragments, and molecules containing the same canonical fragment share a fragment node. For every fragment $f$, we record the number of indexed molecules containing it as $c_f$; this document frequency is used to reduce the influence of ubiquitous fragments during ranking.
  \item \textbf{Functional group.} We use all 85 functional-group descriptors provided by RDKit rather than selecting patterns manually. Molecules share a functional-group node whenever the corresponding descriptor has a nonzero count, while the count itself is retained for similarity scoring.
  \item \textbf{Fingerprint.} An ECFP4 fingerprint (Morgan radius 2; 2,048 bits) is computed for every molecule. Unlike the other views, a fingerprint supports direct pairwise Tanimoto comparison and therefore does not require an intervening structure node.
\end{itemize}

\subsection{Bridge Index}

The bridge index allows a structurally retrieved molecule to reach biomedical evidence even when the two layers use different identifiers. For every molecular-layer entity, we identify all equivalent bio-annotated entities by applying the following matching rules in order:
\begin{enumerate}[nosep]
  \item \textbf{direct identifier match.} both layers already use the same identifier, as for shared ChEBI entries.
  \item \textbf{declared equivalence.} Records from different namespaces, such as PubChem and DrugBank, are connected via an RTX-KG2 \texttt{biolink:same\_as} relation.
  \item \textbf{InChIKey.} the molecule's SMILES is converted to an InChIKey with RDKit; a bio-annotated entity with the same InChIKey is treated as an equivalent molecular record.
\end{enumerate}
We retain both the matched entity and the rule that established the equivalence. Retrieval and traversal use this mapping before entering the biomedical layer, so an anchor with no biomedical relations of its own can still reach evidence attached to an equivalent record. Because the bridge is maintained separately from both layers, either layer can be replaced or updated without changing the representation of the other.

\section{In-Graph Link Recovery: Benchmark Construction}
\label{app:tasks}

This benchmark evaluates whether structural anchors can recover biomedical relations that have been hidden from otherwise known drug entities. It separates the definition of each task from the masking protocol and the generalization regime, making clear which evidence remains available at evaluation time.

\subsection{Tasks and Candidate Spaces}
Table~\ref{tab:app_tasks} specifies the predicates, evidence sources, and candidate spaces used in the four masked-edge retrieval tasks. T1 and T2 use source-restricted edges from DrugCentral; T3 uses source-restricted edges from DrugBank, ChEMBL, and DrugCentral; T4 uses directional \texttt{affects} edges from the frozen RTX-KG2c snapshot.

\begin{table*}[t]
\centering
\caption{Masked-edge task definitions. T4 candidates are composite labels $\langle\text{object}\mid\text{direction}\rangle$, where direction is \texttt{increased} or \texttt{decreased}.}
\label{tab:app_tasks}
\resizebox{\textwidth}{!}{%
\begin{tabular}{lllll}
\toprule
Task & Evaluated relation & Gold-edge source & Candidate space & Regimes \\
\midrule
T1: Indication
  & \texttt{treats\_or\_applied\_or\_studied\_to\_treat}
  & DrugCentral & Disease and phenotype nodes & L1--L4 \\
T2: Contraindication
  & \texttt{contraindicated\_in}
  & DrugCentral & Objects of \texttt{contraindicated\_in} & L1--L4 \\
T3: Drug--target
  & \texttt{affects}, \texttt{physically\_interacts\_with},
    \texttt{directly\_physically\_interacts\_with}
  & DrugBank, ChEMBL, DrugCentral & Gene and protein nodes & L1--L3 \\
T4: Directional effect
  & \texttt{affects} with a direction annotation
  & RTX-KG2c & 58,261 object--direction pairs & L1--L3 \\
\bottomrule
\end{tabular}%
}
\end{table*}

\subsection{Masking and Evaluation Protocol}

For each task and generalization regime, evaluation proceeds as follows:
\begin{enumerate}
  \item Drugs are partitioned into training and test sets according to the regime defined below.
  \item For every test drug, the task-relevant outgoing biomedical edges are removed from the evaluation graph. The corresponding edges of training drugs remain available as retrieval evidence.
  \item The query drug is mapped to structural anchors among the unmasked molecules. Candidate endpoints are collected only through the anchors' retained biomedical relations; the held-out endpoints of the query drug are not directly accessible.
  \item The ranking pool contains the gold endpoints and 99 negatives sampled from the task-specific candidate space. Negative sampling is seeded by the query-drug identifier and is therefore deterministic across methods.
\end{enumerate}

The reported metrics are Hits@1, Hits@10, and mean reciprocal rank (MRR). Query molecules are excluded from their own anchor sets by canonical-SMILES and InChIKey matching.

\subsection{Generalization Regimes}

The four regimes test different forms of generalization. L1 measures performance under a random split, L2 and L3 introduce increasingly different structural neighborhoods, and L4 evaluates transfer across disease areas rather than chemical novelty.

\begin{itemize}
  \item \textbf{L1 (random):} drugs are randomly split into 80\% training and 20\% test sets. Close structural neighbors may remain across the split.
  \item \textbf{L2 (scaffold holdout):} complete Bemis--Murcko scaffold groups are assigned to one split, so a test drug cannot share its exact scaffold with a training drug.
  \item \textbf{L3 (ECFP4-cluster holdout):} drugs are grouped by greedy Butina clustering over ECFP4 similarities at threshold $\tau=0.4$, and complete clusters are assigned to one split. This creates a fingerprint-level distribution shift, but does not impose scaffold disjointness or a strict pairwise similarity bound on every training--test pair.
  \item \textbf{L4 (disease-area holdout; T1/T2 only):} complete MONDO disease areas are withheld. L4 measures transfer across therapeutic areas rather than structural novelty.
\end{itemize}

\section{Masked-Edge Coverage and Complete Results}
\label{app:masked_results}

\subsection{Evaluation-Set Statistics}

Table~\ref{tab:app_stats} reports the complete evaluatable test-set size and gold-label counts. The TxGNN comparison in the main paper is restricted to drugs represented in the TxGNN vocabulary; Table~\ref{tab:app_coverage} reports this overlap for L1.

\begin{table*}[t]
\centering
\caption{Complete test-set statistics for the masked-edge benchmark. ``Total gold'' is the sum of the per-drug gold-set sizes.}
\label{tab:app_stats}
\small
\begin{tabular}{ll rrr}
\toprule
Task & Regime & Test drugs & Total gold & Median gold per drug \\
\midrule
\multirow{4}{*}{T1: Indication}
  & L1 & 2,097 & 44,143 & 3 \\
  & L2 & 2,106 & 59,897 & 3 \\
  & L3 & 2,096 & 66,478 & 4 \\
  & L4 & 2,096 & 57,936 & 5 \\
\midrule
\multirow{4}{*}{T2: Contraindication}
  & L1 & 260 & 4,286 & 10 \\
  & L2 & 264 & 6,333 & 16 \\
  & L3 & 260 & 5,572 & 12 \\
  & L4 & 260 & 6,578 & 18 \\
\midrule
\multirow{3}{*}{T3: Drug--target}
  & L1 & 4,315 & 112,753 & 3 \\
  & L2 & 4,318 & 162,767 & 3 \\
  & L3 & 4,315 & 198,222 & 4 \\
\midrule
\multirow{3}{*}{T4: Directional effect}
  & L1 & 3,738 & 109,889 & 3 \\
  & L2 & 3,751 & 158,440 & 4 \\
  & L3 & 3,748 & 192,298 & 4 \\
\bottomrule
\end{tabular}%
\end{table*}

\begin{table}[t]
\centering
\caption{TxGNN vocabulary coverage for the L1 test sets. The main-paper comparison evaluates both methods on the overlap subset.}
\label{tab:app_coverage}
\small
\begin{tabular}{lrrr}
\toprule
Task & Complete set & Overlap & Coverage \\
\midrule
T1 & 2,097 & 312 & 14.9\% \\
T2 & 260   & 154 & 59.2\% \\
T3 & 4,315 & 366 & 8.5\% \\
T4 & 3,738 & 333 & 8.9\% \\
\bottomrule
\end{tabular}
\end{table}

\subsection{Complete-Set Performance}

Table~\ref{tab:app_full_metrics} reports \molbiokg on every evaluatable drug, without restricting the test set to the TxGNN vocabulary. These values are complementary to the controlled overlap comparison in the main paper and should not be compared directly across different test-set scopes.

\begin{table*}[t]
\centering
\caption{\molbiokg performance on the complete evaluatable test sets ($K=20$ anchors per family, 99 sampled negatives). L4 does not apply to T3 or T4.}
\label{tab:app_full_metrics}
\small
\begin{tabular}{ll rrr}
\toprule
Task & Regime & H@1 & H@10 & MRR \\
\midrule
\multirow{4}{*}{T1: Indication}
  & L1 & .355 & .601 & .453 \\
  & L2 & .131 & .430 & .246 \\
  & L3 & .313 & .601 & .420 \\
  & L4 & .312 & .551 & .413 \\
\midrule
\multirow{4}{*}{T2: Contraindication}
  & L1 & .385 & .765 & .503 \\
  & L2 & .224 & .701 & .362 \\
  & L3 & .304 & .689 & .424 \\
  & L4 & .465 & .850 & .589 \\
\midrule
\multirow{3}{*}{T3: Drug--target}
  & L1 & .386 & .610 & .465 \\
  & L2 & .114 & .521 & .246 \\
  & L3 & .303 & .612 & .408 \\
\midrule
\multirow{3}{*}{T4: Directional effect}
  & L1 & .373 & .640 & .464 \\
  & L2 & .127 & .584 & .275 \\
  & L3 & .316 & .661 & .430 \\
\bottomrule
\end{tabular}
\end{table*}

Table~\ref{tab:app_scope_l1} shows the effect of evaluation scope at L1. The overlap rows reproduce the controlled comparison in the main paper; the complete rows evaluate \molbiokg on all available test drugs.

\begin{table*}[t]
\centering
\caption{L1 results on the TxGNN-covered overlap and on the complete \molbiokg test set.}
\label{tab:app_scope_l1}
\small
\begin{tabular}{lll rrr}
\toprule
Task & Scope & Method & H@1 & H@10 & MRR \\
\midrule
\multirow{3}{*}{T1}
  & \multirow{2}{*}{Overlap} & TxGNN & .542 & .843 & .656 \\
  & & \molbiokg & .660 & .885 & .750 \\
  & Complete & \molbiokg & .355 & .601 & .453 \\
\midrule
\multirow{3}{*}{T2}
  & \multirow{2}{*}{Overlap} & TxGNN & .604 & .727 & .670 \\
  & & \molbiokg & .416 & .812 & .536 \\
  & Complete & \molbiokg & .385 & .765 & .503 \\
\midrule
\multirow{3}{*}{T3}
  & \multirow{2}{*}{Overlap} & TxGNN & .235 & .757 & .402 \\
  & & \molbiokg & .710 & .855 & .763 \\
  & Complete & \molbiokg & .386 & .610 & .465 \\
\midrule
\multirow{3}{*}{T4}
  & \multirow{2}{*}{Overlap} & TxGNN & .285 & .748 & .447 \\
  & & \molbiokg & .739 & .898 & .802 \\
  & Complete & \molbiokg & .373 & .640 & .464 \\
\bottomrule
\end{tabular}
\end{table*}

\section{Retrieval Strategy Ablation}
\label{app:ablation_full}

This ablation tests whether the four structural views provide complementary evidence or merely duplicate one another. We fix the retrieval budget at $K=5$ for every enabled family, ensuring that adding a family changes the type of evidence available without increasing the budget assigned to any individual family. Table~\ref{tab:app_ablation_full} provides the values underlying the main-paper ablation figure.

\begin{table*}[t]
\centering
\caption{T1 anchor-strategy ablation on the complete L2 and L3 test sets ($K=5$ per enabled family). Multi-family rankings are combined by reciprocal rank fusion. \textbf{Bold}: best result in each column.}
\label{tab:app_ablation_full}
\small
\begin{tabular}{l rrr rrr}
\toprule
& \multicolumn{3}{c}{L2: Scaffold holdout} & \multicolumn{3}{c}{L3: ECFP4-cluster holdout} \\
\cmidrule(lr){2-4}\cmidrule(lr){5-7}
Strategy & H@1 & H@10 & MRR & H@1 & H@10 & MRR \\
\midrule
Scaffold only & .131 & .430 & .246 & .221 & .528 & .338 \\
BRICS fragment only & .185 & .463 & .291 & .176 & .487 & .292 \\
Functional group only & .343 & .592 & .438 & .311 & .585 & .415 \\
Fingerprint only & .294 & .528 & .387 & .327 & .572 & .419 \\
\midrule
Fingerprint + Scaffold & .294 & .528 & .387 & .370 & .609 & .463 \\
Fingerprint + Scaffold + Fragment & .324 & .548 & .413 & .384 & .617 & .475 \\
All four families & \textbf{.408} & \textbf{.656} & \textbf{.504} & \textbf{.414} & \textbf{.676} & \textbf{.517} \\
\bottomrule
\end{tabular}%
\end{table*}

No individual family is strongest in both regimes. Functional groups provide the highest standalone H@1 at L2, fingerprints provide the highest standalone H@1 at L3, and the four-family combination is best on all reported metrics in both regimes. Under L2, Fingerprint and Fingerprint~+~Scaffold report identical metrics because the Scaffold channel contributes weakly in this regime; adding it to the fused ranking does not change the top-$K$ order for any query, so the RRF output is unaffected.

\section{Static Multi-Anchor Retrieval Details}
\label{app:retrieval}

Static inference consists of two stages. Each structural view first ranks graph-resident molecules according to its own notion of similarity; Reciprocal Rank Fusion (RRF) then combines these rankings before biomedical evidence is collected. The definitions below provide the scoring and retrieval details omitted from the main paper.

\subsection{Anchor Extraction and Ranking}

\begin{enumerate}
  \item \textbf{Scaffold.} The query's Bemis--Murcko scaffold is matched against the scaffold index. If no indexed molecule shares the exact scaffold, retrieval falls back to the scaffold's generic form -- its ring/connectivity skeleton with all heteroatoms replaced by carbon and all bonds set to single bonds -- at half the exact-match score to reflect the coarser match.
  \item \textbf{BRICS fragment.} The query is decomposed by BRICS. For query fragments $F_q$ and candidate fragments $F_a$, the score is
  \begin{equation}
    s_{\mathrm{frag}}(a)=
    \frac{\sum_{f\in F_q\cap F_a}\mathrm{idf}(f)}
         {\sum_{f\in F_q}\mathrm{idf}(f)},
  \end{equation}
  where $\mathrm{idf}(f)=\log((N+1)/(c_f+1))+1$ and $c_f$ is the number of indexed molecules containing fragment $f$.
  \item \textbf{Functional group.} Each molecule is represented as an 85-dimensional vector over RDKit's \texttt{fr\_*} descriptors, where entry $f$ holds $\mathrm{idf}(f)\cdot\mathrm{count}_f(a)$ rather than a binary indicator of whether the descriptor is active. Candidates are ranked by cosine similarity between the query's and the candidate's vectors.
  \item \textbf{Fingerprint.} ECFP4 fingerprints use radius 2 and 2,048 bits. A popcount-ratio filter with threshold $0.3$ -- retaining only candidates whose bit count lies within $[0.3, 1/0.3]$ of the query's, the necessary range for Tanimoto similarity of at least $0.3$ -- removes clearly dissimilar candidates before the remaining molecules are ranked by exact Tanimoto similarity (see Supplementary Section~\ref{app:bio_degree} for the bio-degree-adjusted re-ranking applied within this pool).
\end{enumerate}

\subsection{Role of Biomedical Connectivity (bio-degree) in Ranking}
\label{app:bio_degree}

In addition to the four structural scores above, every molecule node carries a \emph{bio-degree} attribute: the number of edges incident to it in the biomedical (Layer-2) KG, summed across all Layer-2 sources. bio-degree is not a fifth structural view and never contributes to the fused RRF ranking directly; instead, it is consulted \emph{within} individual channels to decide between candidates, and its role differs qualitatively across channels rather than uniformly acting as a tie-break.

\begin{itemize}
  \item \textbf{Scaffold.} Every candidate sharing a given (generic) scaffold receives an identical scaffold score, so the score itself cannot distinguish which candidates fill the top-$K$ slots for this channel. bio-degree, taken in descending order, is the criterion that does: it is not a tie-break layered on top of an already-differentiated ranking, but the sole ordering signal within a shared-scaffold group.
  \item \textbf{BRICS fragment.} Candidates are primarily ordered by the IDF-weighted fragment score $s_{\mathrm{frag}}$ (Equation~2); bio-degree is consulted only as a secondary key to break exact ties in $s_{\mathrm{frag}}$, which occur when two candidates match an identical multiset of query fragments.
  \item \textbf{Functional group.} The cosine-similarity ranking over 85-dimensional FG vectors does not consult bio-degree at any stage.
  \item \textbf{Fingerprint.} Within the Tanimoto-filtered candidate pool, candidates are ordered by $s_{\mathrm{tan}}(a)\cdot(1+\log(1+\mathrm{bio-degree}(a)))$ rather than by $s_{\mathrm{tan}}(a)$ alone. This is a continuous multiplicative re-ranking, not a tie-break: it can change the relative order of two candidates whose Tanimoto scores differ. The value passed on to Reciprocal Rank Fusion is the raw Tanimoto score, but the position each candidate occupies in the list RRF consumes is determined by the bio-degree-adjusted order, so the effect still propagates into the fused ranking.
\end{itemize}


\subsection{Rank Fusion and Evidence Collection}

For a candidate anchor $a$, reciprocal rank fusion combines the available anchor rankings:
\begin{equation}
  \mathrm{RRF}(a)=\sum_{r\in\mathcal{R}}
  \frac{1}{c+\operatorname{rank}_{r}(a)},
\end{equation}
where $\mathcal{R}$ is the set of enabled retrieval families and $c$ is held fixed across experiments. Each family contributes at most $K$ candidates before deduplication. The masked-edge experiments use $K=20$; the controlled strategy ablation uses $K=5$.

After fusion, the top anchors are traversed through identifier-equivalence edges into the biomedical layer. The system retains the anchor identity, structural score, traversed predicate, endpoint, and source record for every collected candidate.

\section{Adapt-KG Prompts and Tool Definitions}
\label{app:adaptkg_prompts}

Adapt-KG is the adaptive counterpart to static RRF. It uses the same iterative decision process for molecule annotation and multi-hop question answering, while the prompt, available actions, and turn budget reflect the needs of each task. At every turn, the model selects an action in a structured format, observes the returned evidence, and either continues gathering evidence or produces its final answer. This section records the prompts and action definitions used in our experiments.

\subsection{Out-of-Graph Generalization}

System prompt:
\begin{quote}
\ttfamily\small
"You are a drug discovery expert. Predict biomedical properties of novel molecules based on SMILES and structural context. Always respond with valid JSON."
\end{quote}

Tool menu (at most 4 tool calls per drug, plus a final answer turn):
\begin{quote}
\ttfamily\small\noindent
Available structural retrieval tools (call one per turn):\\[2pt]
\phantom{x}\ \ scaffold\_search\ \ \ \ ---\ retrieve drugs sharing the same Bemis-Murcko scaffold\\
\phantom{x}\ \ fingerprint\_search --- retrieve drugs with high ECFP4 Tanimoto similarity\\
\phantom{x}\ \ fragment\_search\ \ \ \ --- retrieve drugs sharing BRICS retrosynthetic fragments\\
\phantom{x}\ \ fg\_search\ \ \ \ \ \ \ \ \ \ --- retrieve drugs sharing pharmacophoric functional groups\\[4pt]
Each tool returns the top-$K$ most similar known drugs with their targets and indications from \molbiokg.\\[4pt]
To call a tool, respond with JSON:\\
\{"action": "\textless tool\_name\textgreater", "reasoning": "\textless why this tool\textgreater"\}\\[4pt]
When you have enough evidence, provide your final answer (at most 5 targets, 5 indications):\\
\{"action": "answer", "targets": ["protein1", ...], "indications": ["disease1", ...], "reasoning": "..."\}
\end{quote}

The four actions correspond to the fixed retrieval strategies reported in the main paper: Fingerprint, Scaffold, BRICS Fragment, and Functional Group. Adapt-KG determines which views to consult, their order, and when sufficient evidence has been gathered. Rankings from the selected views are combined using the same RRF procedure as the static method (Section~\ref{app:retrieval}). To keep evaluation comparable across strategies, every final answer is limited to five targets and five indications.

\subsection{Complex Multi-Hop Reasoning}

System prompt:
\begin{quote}
\ttfamily\small
"You are a biomedical knowledge graph reasoning agent. Answer multi-hop questions by calling KG tools to gather evidence. Plan your traversal strategically --- only call tools that advance toward the answer. Stop as soon as you have sufficient evidence."
\end{quote}

Tool menu (at most 6 turns total, including the final answer):
\begin{quote}
\ttfamily\small\noindent
Available KG tools (call one per turn):\\[4pt]
hop\_drug\_disease(entity\_id, direction)\\
\phantom{xx}direction="reverse": disease $\rightarrow$ drugs that treat it\\
\phantom{xx}direction="forward": drug $\rightarrow$ diseases it treats\\[4pt]
hop\_drug\_target(entity\_id, direction)\\
\phantom{xx}direction="reverse": target/protein $\rightarrow$ drugs acting on it\\
\phantom{xx}direction="forward": drug $\rightarrow$ its protein targets\\[4pt]
hop\_disease\_phenotype(entity\_id)\\
\phantom{xx}disease $\rightarrow$ associated clinical phenotypes\\[4pt]
hop\_ppi(entity\_id)\\
\phantom{xx}protein $\rightarrow$ protein-protein interaction partners\\[4pt]
hop\_gene\_disease(entity\_id)\\
\phantom{xx}gene/protein $\rightarrow$ diseases it is genetically associated with\\[4pt]
hop\_mol\_fg\_of\_drug(entity\_id)\\
\phantom{xx}drug $\rightarrow$ its pharmacophoric functional groups (top-10 by frequency)\\[4pt]
hop\_mol\_drugs\_with\_fg(entity\_id)\\
\phantom{xx}functional group ID/name $\rightarrow$ drugs containing it\\[4pt]
hop\_moa\_direction(entity\_id, direction)\\
\phantom{xx}direction="reverse": target $\rightarrow$ drugs that modulate it (with direction label)\\[4pt]
hop\_fg\_aggregate(entity\_id)\\
\phantom{xx}disease $\rightarrow$ top-10 most common functional groups across ALL drugs treating it\\
\phantom{xx}(batch COUNT across all treating drugs --- use for "most prevalent FG" questions)\\[4pt]
To call a tool, respond with JSON:\\
\{"action": "\textless tool\_name\textgreater", "entity\_id": "\textless CURIE or name\textgreater", "direction": "\textless optional\textgreater", "reasoning": "\textless why\textgreater"\}\\[4pt]
When you have enough evidence, respond with:\\
\{"action": "answer", "entities": ["name1", "name2", ...], "reasoning": "\textless 1-2 sentences\textgreater"\}
\end{quote}

These nine actions implement the typed traversals described in the main paper's \textit{Adaptive Retrieval and KG Traversal} subsection. Each action corresponds to a defined transition in the molecular or biomedical layer. For example, \texttt{hop\_ppi} retrieves protein-interaction partners, whereas \texttt{hop\_fg\_aggregate} identifies the most frequent functional groups across all drugs treating a disease. The model is instructed to preserve the entity names returned by the tools rather than paraphrasing them, ensuring that predicted entities can be matched consistently to gold names and aliases.

\section{Baseline Algorithms for Multi-Hop QA}
\label{app:baselines}

The main paper compares six policies for using the same typed traversal actions on MolBioKG-KGQA. This controlled design isolates how each policy selects, organizes, and composes evidence: every method begins with only the question and a topic entity, and no method receives a precomputed answer context. The five comparison policies are described below; Adapt-KG is specified in Section~\ref{app:adaptkg_prompts}. Per-question inference cost is reported in Section~\ref{app:kgqa_compute}.

\paragraph{Zero-shot.} No tool access. The model receives only the question and answers directly from its parametric knowledge in a single call.

\paragraph{RAG (flat retrieval).} This policy tests whether a small amount of unstructured retrieval is sufficient. It first selects one to three traversals from the question and starting entity, then executes them and presents up to 60 retrieved entities as flat text. A second model call answers from this context.

\paragraph{Graph-RAG.} This policy uses the same plan-then-answer structure as RAG but preserves graph organization. The selected traversals are executed jointly, so aggregate operations remain grouped rather than being expanded into unrelated per-drug records. Retrieved entities are organized by type and accompanied by ranks or counts before the model produces its answer.

\paragraph{Graph-CoT.} This policy traverses the graph one step at a time. At each step, the model observes the current entity type, available transitions, and a short reasoning trace, then either selects one transition or stops. When more than 15 entities are returned, an additional selection step retains the most relevant candidates. The process continues for at most five hops before answering from the retained entities and accumulated trace.

\paragraph{Think-on-Graph.} This policy performs beam search over candidate entities. At each of at most five iterations, the model scores the available transitions, executes the highest-scoring one, and ranks the resulting entities by relevance to the question. The five highest-ranked entities continue to the next iteration. A final model call answers from the traversal trace and the surviving candidates.

\section{Inference Cost on Out-of-Graph Annotation}
\label{app:compute}

This analysis examines the computational trade-off between fixed retrieval and adaptive evidence gathering. Table~\ref{tab:app_compute_oog} reports retrieval actions, cumulative token counts, per-drug elapsed time, and Rec\textsubscript{llm} for gpt-oss-120b. Fixed strategies perform one structural retrieval step followed by answer synthesis, whereas Adapt-KG may gather evidence over several turns.

Retrieval actions, token counts, and Rec\textsubscript{llm} are reported as mean$\pm$std over the same three runs on all 199 out-of-graph drugs used in the main paper. Elapsed time is derived from each strategy's own wall-clock pass over all 199 drugs, recorded once per run and averaged over the three runs; because all strategies were measured under shared computational conditions rather than in isolated runs, it indicates practical throughput but should not be interpreted as a controlled latency comparison.

\begin{table*}[t]
\centering
\caption{Per-drug inference cost and recall on the 199-drug OOG benchmark, gpt-oss-120b. \textbf{Bold}: best mean per column among retrieval-based strategies.}
\label{tab:app_compute_oog}
\small
\begin{tabular}{l rrr r rr}
\toprule
Method & Retrieval actions & Input tokens & Output tokens & Time/drug (s) & Ind.\ Rec\textsubscript{llm} & Tgt.\ Rec\textsubscript{llm} \\
\midrule
Zero-shot & 0 & $336\pm0$ & $927\pm17$ & 5.75$\pm$0.55 & 0.186$\pm$0.006 & 0.145$\pm$0.016 \\
\midrule
Fingerprint & 1 & $1{,}111\pm0$ & $726\pm13$ & 6.00$\pm$0.52 & 0.221$\pm$0.013 & 0.208$\pm$0.009 \\
Scaffold & 1 & $438\pm0$ & $803\pm11$ & 7.14$\pm$0.62 & 0.208$\pm$0.009 & 0.186$\pm$0.009 \\
BRICS fragment & 1 & $1{,}223\pm0$ & $790\pm5$ & \textbf{5.65$\pm$0.76} & 0.215$\pm$0.006 & 0.219$\pm$0.010 \\
Functional group & 1 & $1{,}076\pm0$ & $696\pm11$ & 6.32$\pm$1.93 & 0.223$\pm$0.003 & 0.212$\pm$0.012 \\
RRF & 1 & $1{,}145\pm0$ & $748\pm4$ & 8.47$\pm$0.59 & \textbf{0.239$\pm$0.005} & \textbf{0.269$\pm$0.009} \\
\midrule
Adapt-KG & $1.89\pm0.18$ & $5{,}340\pm500$ & $1{,}986\pm33$ & 12.02$\pm$0.62 & 0.232$\pm$0.009 & 0.218$\pm$0.015 \\
\bottomrule
\end{tabular}%
\end{table*}

Among the single-anchor strategies, Scaffold uses the smallest mean input context, while RRF attains the highest recall on both axes, ahead of every single-anchor view; among the single-anchor strategies alone, BRICS Fragment gives the highest target recall. These results show that static fusion provides a strong accuracy--cost trade-off for focused molecular annotation.

Adapt-KG performs 1.89 retrieval actions on average and uses approximately $4.7\times$ as many input tokens as RRF. On this single-hop task, the additional adaptive budget does not improve recall: indication recall remains below RRF, and target recall remains below both RRF and BRICS Fragment, though still above Scaffold, Fingerprint, and Functional Group. BRICS Fragment has the lowest observed per-drug elapsed time among the retrieval-based strategies, while RRF and Adapt-KG are the slowest; the timing differences should nevertheless be interpreted under the caveat above.

\section{MolBioKG-KGQA Benchmark Details}
\label{app:kgqa_construction}

MolBioKG-KGQA contains 460 questions instantiated from graph paths in the frozen \molbiokg snapshot. Table~\ref{tab:app_kgqa_templates} summarizes the five template families. Gold answers are the terminal entities reached by the instantiated paths; the benchmark therefore measures access to and composition of graph-resident evidence rather than external biomedical validity.

\begin{table*}[t]
\centering
\caption{MolBioKG-KGQA template families. Arrows summarize the entity sequence used to instantiate each question family.}
\label{tab:app_kgqa_templates}
\small
\resizebox{\textwidth}{!}{%
\begin{tabular}{lrrl}
\toprule
Template family & Instances & Hops & Path pattern \\
\midrule
Disease to target & 100 & 2 & disease $\rightarrow$ drug $\rightarrow$ target \\
Target to disease & 100 & 2 & target $\rightarrow$ drug $\rightarrow$ disease \\
Disease to functional group & 100 & 2 & disease $\rightarrow$ drug $\rightarrow$ functional group \\
Target modulation to functional group & 100 & 2 & target $\rightarrow$ direction-specific drug $\rightarrow$ functional group \\
Functional group to phenotype & 60 & 5 & functional group $\rightarrow$ drug $\rightarrow$ target $\rightarrow$ interacting protein $\rightarrow$ disease $\rightarrow$ phenotype \\
\bottomrule
\end{tabular}%
}
\end{table*}

\subsection{Question Paraphrase Templates}

For each instance, question text is sampled uniformly from five human-authored paraphrases of its template family, with placeholders replaced by the sampled anchor entity's name. The target-modulation family uses separate paraphrase sets for mechanism-of-action relations annotated as \emph{decreased} and \emph{increased}.

\paragraph{Disease to target.}
Gold answers for this family are capped to the ten most frequent protein targets (TF-IDF-ranked over the disease's associated drugs, downweighting hub targets) rather than the full target set, and question phrasing is worded accordingly.
\begin{enumerate}[nosep]
\item What are the most common protein targets of drugs used to treat \{disease\}?
\item Which protein targets appear most frequently among approved treatments for \{disease\}?
\item What are the dominant molecular targets of therapeutics for \{disease\}?
\item Which genes or proteins are most frequently bound by drugs indicated for \{disease\}?
\item If you surveyed the target profile of drugs for \{disease\}, which protein targets would dominate?
\end{enumerate}

\paragraph{Target to disease.}
As with disease to target, gold answers are capped to the ten most frequent indications.
\begin{enumerate}[nosep]
\item What are the most common diseases treated by drugs that target \{gene\}?
\item Which therapeutic indications appear most frequently among drugs known to interact with \{gene\}?
\item What are the dominant clinical conditions addressed by compounds that target the protein \{gene\}?
\item Which diseases most commonly have established drug treatments that act on \{gene\}?
\item If you surveyed drugs that modulate \{gene\} activity, which clinical indications would dominate?
\end{enumerate}

\paragraph{Disease to functional group.}
Gold answers for this family are capped to the ten most frequent functional groups by raw occurrence count across the disease's CHEBI-identified associated drugs.
\begin{enumerate}[nosep]
\item What functional groups are most common in drugs used to treat \{disease\}?
\item Which structural motifs characterize the drugs approved for \{disease\}?
\item What chemical features are most prevalent among therapeutics indicated for \{disease\}?
\item Which functional groups appear most frequently in the structures of \{disease\} treatments?
\item If you surveyed the chemical space of \{disease\} drugs, which functional groups would dominate?
\end{enumerate}

\paragraph{Target modulation to functional group --- decreased.}
As with disease to functional group, gold answers are capped to the ten most frequent functional groups by raw count, but here the underlying drug set is first restricted to those whose mechanism-of-action edge to \{gene\} is annotated as \emph{decreased} before functional groups are counted.
\begin{enumerate}[nosep]
\item What structural features are shared by drugs that decrease the activity of \{gene\}?
\item Which functional groups are most common among compounds that inhibit \{gene\}?
\item What chemical motifs characterize known inhibitors of \{gene\}?
\item Which structural elements recur in drugs that downregulate \{gene\} activity?
\item What are the dominant functional groups found in drugs that suppress \{gene\}?
\end{enumerate}

\paragraph{Target modulation to functional group --- increased.}
Constructed identically to the decreased variant, but the underlying drug set is restricted to those whose mechanism-of-action edge to \{gene\} is annotated as \emph{increased}.
\begin{enumerate}[nosep]
\item What structural features are shared by drugs that increase the activity of \{gene\}?
\item Which functional groups are most common among compounds that activate \{gene\}?
\item What chemical motifs characterize known activators of \{gene\}?
\item Which structural elements recur in drugs that upregulate \{gene\} activity?
\item What are the dominant functional groups found in drugs that enhance \{gene\} function?
\end{enumerate}

\paragraph{Functional group to phenotype.}
Unlike the other four families, gold answers here are not frequency-ranked: they are the deduplicated union of all phenotypes reachable via the full 5-hop path (drugs containing the functional group $\rightarrow$ their gene targets $\rightarrow$ PPI partners of those genes $\rightarrow$ diseases associated with the partner genes $\rightarrow$ phenotypes of those diseases), capped at the first 30 encountered rather than the 30 most frequent.
\begin{enumerate}[nosep]
\item What clinical phenotypes are linked to the protein interaction network of targets of drugs containing \{fg\}?
\item Through network medicine, what symptoms manifest from the PPI neighborhood of \{fg\}-drug targets?
\item What phenotypic features are associated with diseases linked to PPI partners of \{fg\}-drug targets?
\item Which clinical features emerge from the disease network of PPI partners of \{fg\}-containing drug targets?
\item What observable phenotypes are genomically connected to the interaction partners of \{fg\}-drug targets?
\end{enumerate}

The main-paper comparison uses the same gpt-oss-120b backbone for all LLM-based methods and reports Hits@1, Hits@5, Hits@10, and entity-set macro-F1 (precision/recall over the predicted vs.\ gold entity sets per instance, averaged across all 460 instances). In the single-layer ablations, a method may access only the selected layer. Because this intervention changes both graph connectivity and the available evidence, the resulting difference should be interpreted as a layer-restriction test rather than a pure structural causal ablation.

\section{Inference Cost on MolBioKG-KGQA}
\label{app:kgqa_compute}

This analysis compares the cost of different evidence-composition policies on the same multi-hop questions. Table~\ref{tab:app_kgqa_compute} reports per-question token use and elapsed time for the methods in the main-paper comparison, all using gpt-oss-120b. Each retrieval-based method begins with only the question and a topic entity, then chooses its own traversal actions; token use therefore reflects the amount of evidence gathered and the number of intermediate decisions.

\begin{table}[t]
\centering
\caption{Per-question inference cost on MolBioKG-KGQA (460 questions, gpt-oss-120b backbone).}
\label{tab:app_kgqa_compute}
\small
\resizebox{\columnwidth}{!}{
\begin{tabular}{lrrr}
\toprule
Method & Input tokens & Output tokens & Time/question (s) \\
\midrule
Zero-shot       & 195.6   & 497.7    & 1.8  \\
RAG            & 1{,}483.5 & 907.6   & 5.7  \\
Think-on-Graph & 2{,}574.5 & 9{,}086.9 & 35.6 \\
Graph-CoT      & 1{,}641.3 & 1{,}781.4 & 7.2  \\
Graph-RAG      & 1{,}342.3 & 785.8   & 8.5  \\
Adapt-KG       & 4{,}692.7 & 1{,}208.9 & 5.0  \\
\bottomrule
\end{tabular}
}
\end{table}

Think-on-Graph's beam search incurs the largest output-token cost because it evaluates intermediate candidates before committing to a path; it also issues the most retrieval actions on average (2.38 vs.\ Adapt-KG's 1.98). Adapt-KG nevertheless consumes the most input context per question, reflecting its practice of gathering broader evidence within each turn, yet it is the fastest tool-using method overall—faster even than flat RAG. Since elapsed time also reflects shared computational conditions beyond the number of model and retrieval steps, it should be read as an indicative efficiency measure rather than a controlled latency comparison.

\section{Out-of-Graph Annotation Benchmark Details}
\label{app:oog_construction}

This benchmark isolates the cold-start setting in which the query molecule provides no direct route to biomedical annotations. Instead of sampling a synthetic holdout, it applies a temporal cutoff: both the molecular and biomedical layers of \molbiokg are restricted to ChEMBL release~10 (June 2011), and evaluation drugs are drawn from compounds first approved afterward.
Because a drug approved after the cutoff cannot appear in the pre-cutoff KG by definition, every evaluation drug is confirmed absent from the 2011 snapshot at both the structural and biomedical level, and out-of-graph status is guaranteed by construction from approval chronology; the per-drug verification below serves as a confirmatory check rather than the primary basis for labeling.

\subsection{Temporal-Cutoff Snapshot}

The pre-cutoff molecular pool is built by parsing every compound in the ChEMBL release~10 chemical-representations file (999,881 records) with RDKit, computing its InChIKey, and matching that key against \molbiokg's identifier index to recover the corresponding main-graph node ID; 949,869 compounds (95.0\%) resolve to a node and form the pre-2011 structural pool. The pre-cutoff biomedical layer is then derived by retaining, in the drug--disease and drug--target subgraphs, only edges whose subject drug node belongs to this pool: 208,325 of 421,939 drug--disease edges are retained (5,734 of 22,470 distinct drug subjects), and 468,814 of 1,127,690 drug--target edges are retained (11,420 of 42,252 distinct drug subjects). No other filtering, such as by evidence source, is applied at this stage; source-level filtering is instead applied at evaluation time, consistent with the masking protocol used for the masked-edge tasks (Supplementary Section~\ref{app:tasks}).

\subsection{Candidate Selection and Out-of-Graph Verification}

Candidate drugs are compounds with a first-approval date in 2012--2024, drawn from ChEMBL mechanism-of-action records, with simple-salt duplicates removed. Each candidate is tagged with an annotation-completeness tier reflecting how many independent indications its ChEMBL record carries: 183 candidates have multiple independent indications and 16 have a single indication and a single target, for 199 candidates in total.

For every candidate, the canonical SMILES is converted to an InChIKey and checked against the pre-2011 structural pool and the pre-2011 drug--disease and drug--target edges retained above. A candidate that already carries a pre-2011 drug--disease edge would not be a genuine out-of-graph case under this snapshot and is dropped rather than kept and mislabeled. Applying this check to all 199 candidates confirms that every candidate's exact structure is absent from the pre-2011 structural pool, none carries a pre-2011 biomedical edge of either type, and none is dropped.

\subsection{Reference Annotations and Metrics}

Reference indications and targets for all 199 drugs are collected from ChEMBL mechanism-of-action records, independent of the KG. Raw ChEMBL target identifiers are resolved to preferred names and HGNC gene symbols, so that exact-match scoring compares against readable names rather than opaque database identifiers. On average, each drug carries 4.0 reference indications and 1.3 reference targets. Exact recall (Rec\textsubscript{ex}) uses normalized surface-form matching and is conservative under synonyms or paraphrases. LLM-judge recall (Rec\textsubscript{llm}) accepts semantically equivalent annotations using a fixed gpt-oss-120b judge applied uniformly to all methods. Overall results are reported as the mean over all 199 out-of-graph drugs.

\section{Out-of-Graph Annotation: Additional Backbones}
\label{app:qwen3}

These experiments test whether the out-of-graph findings persist across LLM backbones. Tables~\ref{tab:app_oog_gemma431bit}--\ref{tab:app_oog_qwen3635ba3b} report results for gemma-4-31b-it and Qwen3.6-35B-A3B on the same 199-drug benchmark, using the same recall definitions, retrieval thresholds, and Random Context control as the main paper. Because generation is stochastic under the sampling settings used here, all LLM-based results are reported as mean$\pm$std over three runs.

\begin{table}[t]
\centering
\caption{Out-of-graph annotation recall for gemma-4-31b-it, reported as mean$\pm$std over 3 runs, using the same evaluation protocol, thresholds, and Random Context control as the gpt-oss-120b table in the main paper. \textbf{Bold}: best mean per column among retrieval-based strategies.}
\label{tab:app_oog_gemma431bit}
\small
\resizebox{\columnwidth}{!}{
\begin{tabular}{l rr rr}
\toprule
& \multicolumn{2}{c}{Exact Recall (Rec\textsubscript{ex})}
& \multicolumn{2}{c}{LLM-Judge Recall (Rec\textsubscript{llm})} \\
\cmidrule(lr){2-3}\cmidrule(lr){4-5}
Strategy & Ind. & Tgt. & Ind. & Tgt. \\
\midrule
Zero-shot (no KG)
  & 0.070$\pm$0.001 & 0.105$\pm$0.004 & 0.163$\pm$0.002 & 0.122$\pm$0.004 \\
Random Context
  & 0.057$\pm$0.000 & 0.081$\pm$0.000 & 0.161$\pm$0.004 & 0.086$\pm$0.000 \\
\midrule
Fingerprint (FP)
  & 0.079$\pm$0.000 & 0.173$\pm$0.000 & 0.192$\pm$0.001 & 0.180$\pm$0.000 \\
Scaffold
  & \textbf{0.083$\pm$0.001} & 0.103$\pm$0.002 & 0.161$\pm$0.001 & 0.127$\pm$0.002 \\
BRICS Fragment
  & 0.077$\pm$0.001 & \textbf{0.224$\pm$0.003} & 0.199$\pm$0.004 & \textbf{0.258$\pm$0.003} \\
Funct.\ Group (FG)
  & 0.081$\pm$0.001 & 0.182$\pm$0.000 & \textbf{0.201$\pm$0.005} & 0.199$\pm$0.000 \\
RRF
  & \textbf{0.083$\pm$0.001} & 0.215$\pm$0.005 & \textbf{0.201$\pm$0.001} & 0.237$\pm$0.005 \\
\midrule
Adapt-KG
  & 0.070$\pm$0.001 & 0.213$\pm$0.003 & 0.199$\pm$0.002 & 0.237$\pm$0.002 \\
\bottomrule
\end{tabular}
}
\end{table}

\begin{table}[t]
\centering
\caption{Out-of-graph annotation recall for Qwen3.6-35B-A3B, reported as mean$\pm$std over 3 runs, using the same evaluation protocol, thresholds, and Random Context control as the gpt-oss-120b table in the main paper. \textbf{Bold}: best mean per column among retrieval-based strategies.}
\label{tab:app_oog_qwen3635ba3b}
\small
\resizebox{\columnwidth}{!}{
\begin{tabular}{l rr rr}
\toprule
& \multicolumn{2}{c}{Exact Recall (Rec\textsubscript{ex})}
& \multicolumn{2}{c}{LLM-Judge Recall (Rec\textsubscript{llm})} \\
\cmidrule(lr){2-3}\cmidrule(lr){4-5}
Strategy & Ind. & Tgt. & Ind. & Tgt. \\
\midrule
Zero-shot (no KG)
  & 0.048$\pm$0.008 & 0.101$\pm$0.028 & 0.052$\pm$0.010 & 0.101$\pm$0.028 \\
Random Context
  & 0.055$\pm$0.005 & 0.073$\pm$0.008 & 0.061$\pm$0.008 & 0.073$\pm$0.008 \\
\midrule
Fingerprint (FP)
  & 0.084$\pm$0.004 & 0.175$\pm$0.007 & 0.096$\pm$0.005 & 0.176$\pm$0.005 \\
Scaffold
  & 0.076$\pm$0.006 & 0.149$\pm$0.008 & 0.086$\pm$0.011 & 0.154$\pm$0.008 \\
BRICS Fragment
  & 0.068$\pm$0.009 & 0.196$\pm$0.004 & 0.083$\pm$0.008 & 0.198$\pm$0.003 \\
Funct.\ Group (FG)
  & 0.064$\pm$0.005 & 0.133$\pm$0.010 & 0.066$\pm$0.006 & 0.133$\pm$0.010 \\
RRF
  & \textbf{0.089$\pm$0.010} & \textbf{0.232$\pm$0.010} & \textbf{0.102$\pm$0.008} & \textbf{0.233$\pm$0.011} \\
\midrule
Adapt-KG
  & 0.085$\pm$0.009 & 0.177$\pm$0.004 & 0.096$\pm$0.006 & 0.177$\pm$0.004 \\
\bottomrule
\end{tabular}
}
\end{table}

Across both additional backbones, most structural retrieval strategies substantially exceed Zero-shot and Random Context under Rec\textsubscript{llm}, reproducing the main conclusion that the gains come from structurally relevant evidence rather than from merely adding a candidate list. The one exception is Scaffold's indication recall on gemma-4-31b-it, which is roughly on par with both baselines even though its target recall still exceeds them. Random Context itself is inconsistent: it sometimes improves indication recall but does not reliably improve target recall.

The strongest strategy varies with the prediction type and backbone. For Qwen3.6-35B-A3B, RRF gives the highest recall on every column. For gemma-4-31b-it, no single view dominates: Scaffold gives the highest exact indication recall, Funct.\ Group the highest LLM-judged indication recall, and BRICS Fragment the highest exact and LLM-judged target recall. Adapt-KG remains competitive but is not the best method for this single-hop annotation task on either backbone. This pattern supports the division of labor observed in the main paper: static retrieval is well suited to focused molecular annotation, while adaptive traversal is most useful when the question requires compositional, multi-hop evidence.

\section{MolBioKG-KGQA: Backbone Robustness}
\label{app:kgqa_backbone}

Table~\ref{tab:app_kgqa_backbone} evaluates the KGQA methods with two additional open-weight backbones, gemma-4-31b-it and Qwen3.6-35B-A3B—the same two backbones used for the out-of-graph annotation robustness check in Section~\ref{app:qwen3}. Dashes denote layer-ablation results that are not applicable to the Zero-shot baseline.

\begin{table}[t]
\centering
\caption{MolBioKG-KGQA results across two additional backbones ($n=460$). Full-KG columns report H@10 and macro-F1. Layer-ablation columns report H@10 with access restricted to the molecular (M) or biomedical (B) layer. \textbf{Bold}: best non-Zero-shot result in each column.}
\label{tab:app_kgqa_backbone}
\small
\setlength{\tabcolsep}{4pt}
\resizebox{\columnwidth}{!}{%
\begin{tabular}{l rr rr @{\hspace{8pt}} rr rr}
\toprule
& \multicolumn{4}{c}{Full KG (H@10 / F1)}
& \multicolumn{4}{c}{Layer-ablation H@10} \\
\cmidrule(lr){2-5}\cmidrule(lr){6-9}
& \multicolumn{2}{c}{gemma-4-31b-it}
& \multicolumn{2}{c}{Qwen3.6-35B-A3B}
& \multicolumn{2}{c}{gemma-4-31b-it}
& \multicolumn{2}{c}{Qwen3.6-35B-A3B} \\
\cmidrule(lr){2-3}\cmidrule(lr){4-5}\cmidrule(lr){6-7}\cmidrule(lr){8-9}
Method & H@10 & F1 & H@10 & F1 & M & B & M & B \\
\midrule
Zero-shot
  & .135 & .019 & .111 & .016
  & --- & --- & --- & --- \\
\midrule
RAG
  & .585 & .420 & \textbf{.872} & \textbf{.641}
  & .000 & .420 & .074 & .415 \\
Think-on-Graph
  & .389 & .158 & .389 & .158
  & .048 & .183 & .030 & .183 \\
Graph-CoT
  & .526 & .332 & .607 & .332
  & \textbf{.054} & .235 & \textbf{.161} & .341 \\
Graph-RAG
  & .530 & .245 & .717 & .473
  & .000 & .400 & .002 & .383 \\
Adapt-KG
  & \textbf{.870} & \textbf{.596} & \textbf{.872} & .612
  & .000 & \textbf{.430} & .120 & \textbf{.526} \\
\bottomrule
\end{tabular}%
}
\end{table}

The best H@10 method again depends on the backbone: Adapt-KG leads outright with gemma-4-31b-it (.870), while Adapt-KG and flat RAG are effectively tied for the best H@10 with Qwen3.6-35B-A3B (.872 each), though RAG attains the higher macro-F1 on that backbone (.641 vs.\ .612). Under Mol-only restriction, Graph-CoT is the most robust policy for both backbones (.054 and .161)—echoing the gpt-oss-120b result in the main paper, where Think-on-Graph rather than Adapt-KG led the Mol-only column. Under Bio-only restriction, Adapt-KG remains strongest for both backbones (.430 and .526). Taken together with the main-paper table, this indicates that Adapt-KG's advantage is most reliable under full-graph and biomedical-only access, while single-layer (Mol-only) robustness is backbone- and context-dependent.

\section{Out-of-Graph Annotation: Disease Synonym and Hierarchy Normalization}
\label{app:oog_synonym}

Exact surface-form matching can underestimate indication recall when a correct prediction uses a synonym or differs in disease granularity from the gold label. This issue is more acute for the temporal-cutoff out-of-graph benchmark than for a hand-curated one: its 342 distinct gold indication names come directly from ChEMBL mechanism-of-action records as free text, with no ontology identifier attached at all. Ontology-based normalization therefore first requires linking each name to a MONDO, EFO, or HP class before any synonym or hierarchy expansion is possible. We perform this linking and re-evaluate the existing RRF predictions from the three gpt-oss-120b runs reported in the main paper, without generating new model outputs.

\paragraph{Name linking.} Each gold indication name is matched against every live MONDO class's preferred label and exact synonyms; a name matching exactly one class is linked to it. Names with no MONDO match are then checked the same way against EFO and then HP. A name matching more than one class under this criterion is left unresolved rather than guessed, since an incorrect link would silently bias the downstream tiers.

\paragraph{Matching tiers.} Once a gold name is linked, we construct cumulative matching tiers from its ontology class. The \emph{exact} tier adds the class's preferred label and exact synonyms, covering renamed terms such as the gold name ``primary biliary cirrhosis'' linking to MONDO's current label ``primary biliary cholangitis.'' The \emph{related} tier additionally accepts related synonyms. The \emph{subtype} tier accepts a more specific prediction when it lies within two descendant relations of a broader gold concept: for example, Adagrasib's gold indications ``lung cancer'' and ``colorectal cancer'' are credited because the model's answer names the more specific ``non-small cell lung carcinoma'' and ``colorectal carcinoma,'' both within two descendant relations of the respective gold terms. Finally, the \emph{supertype} tier also accepts ancestors within two relations: Vorasidenib's gold indications ``astrocytoma'' and ``low grade glioma'' are credited because the model instead answers with the broader ``glioma.''

\paragraph{Ontology alignment and safeguards.} Of the 342 distinct gold indication names, 291 are linked through an exact label or exact-synonym match in MONDO and 33 more through EFO or HP; 5 match more than one class and are left unresolved, and 13 have no exact match in any of the three ontologies, typically because ChEMBL records the name in an inverted ``Term, Qualifier'' order (e.g.\ ``Hemoglobinuria, Paroxysmal'') that does not appear verbatim in any ontology label or synonym. To avoid giving excessive credit for broad disease concepts, subtype matching is disabled when a gold term has more than 50 descendants within two relations; this safeguard applies to 24 concepts, including ``neoplasm,'' ``carcinoma,'' and ``cancer.''

\paragraph{Scope.} We apply normalization only to indications. Target naming needs no such step here: all 264 gold target references in the benchmark already resolve to a real protein name, and 248 (94\%) additionally carry gene-symbol aliases (Section~\ref{app:oog_construction}).

\begin{table}[t]
\centering
\caption{RRF indication recall under cumulative synonym and hierarchy matching, gpt-oss-120b, mean$\pm$std over the same 3 runs used throughout the paper. \emph{Original} reproduces the main-paper Rec\textsubscript{ex} result; each subsequent tier includes the preceding tiers.}
\label{tab:app_oog_synonym_recall}
\small
\begin{tabular}{lr}
\toprule
Tier & Overall \\
\midrule
Original (main paper) & 0.125$\pm$0.003 \\
+ exact synonym      & 0.181$\pm$0.004 \\
+ related synonym    & 0.184$\pm$0.003 \\
+ subtype (headline) & \textbf{0.210$\pm$0.003} \\
+ supertype          & 0.305$\pm$0.005 \\
\bottomrule
\end{tabular}
\end{table}

\begin{table}[t]
\centering
\caption{Attribution of RRF's 135 indication-axis failures (of 199 out-of-graph drugs, one representative gpt-oss-120b run) to the first tier that correctly matches each case. \emph{Unresolved} cases are not explained by naming or granularity mismatch at any tier.}
\label{tab:app_oog_synonym_attribution}
\small
\begin{tabular}{lr}
\toprule
Lowest rescuing tier & Count \\
\midrule
Exact synonym   & 12 \\
Related synonym & 1 \\
Subtype         & 8 \\
Supertype       & 21 \\
\midrule
Unresolved (genuine miss) & 93 \\
\midrule
Total failures  & 135 \\
\bottomrule
\end{tabular}
\end{table}

Table~\ref{tab:app_oog_synonym_recall} shows that exact-synonym and subtype matching raises RRF indication recall from 0.125 to 0.210, an increase of 8.5 points or 68\% relative. We treat this cumulative tier as the primary normalized result because it credits a more specific correct answer without accepting a less specific one. Adding supertype matching raises recall further to 0.305 but uses a more permissive criterion.

Table~\ref{tab:app_oog_synonym_attribution} explains the 135 indication failures under the original exact-match evaluation, for one representative run; the other two runs show a consistent pattern (133 and 134 failures, with a similar tier split). Exact and related synonyms account for 13 cases, subtype granularity for 8, and supertype matching for 21 more --- 42 failures in total, or 31\%, are naming or granularity artifacts rather than retrieval or generation misses. The remaining 93 cases, or 69\%, are not explained by naming or hierarchy differences at any tier.

\section{Case Studies: Anchor-Family Complementarity}
\label{app:oog_qual_structures}

These cases illustrate why the four structural views are complementary. Each example follows the same evidence chain---query$\to$structural anchor$\to$retrieved analogue$\to$KG relation$\to$candidate answer---and identifies both the view that loses relevant information and the view that recovers it. Cases C--G come from the main out-of-graph benchmark. Cases A and B use a molecular collection restricted to records available before 2011, with queries standardized to their principal molecular component. This design prevents an apparently successful retrieval from being caused by an equivalent salt representation of the query itself. Table~\ref{tab:app_case_summary} provides the high-level pattern before the individual evidence paths are examined.

\begin{table*}[t]
\centering
\caption{Summary of the seven qualitative cases. Each row identifies the limitation exposed by one structural view and the complementary view that retrieves relevant biomedical evidence.}
\label{tab:app_case_summary}
\small
\resizebox{\textwidth}{!}{%
\begin{tabular}{clll}
\toprule
Case & Query & Limitation exposed & Complementary evidence \\
\midrule
A & Binimetinib & Functional groups miss the close analogue & Scaffold retrieves Selumetinib \\
B & Ensartinib & No whole-scaffold match & BRICS retrieves Crizotinib \\
C & Vanzacaftor & Local substructure views find no useful anchor & Fingerprint retrieves Elexacaftor \\
D & Landiolol Hydrochloride & Fragment matches lack the gold indication & Fingerprint retrieves Xamoterol \\
E & Gepirone Hydrochloride & Functional groups are too nonspecific & BRICS and Fingerprint retrieve azapirones \\
F & Lazertinib Mesylate & Relevant analogues rank low by Fingerprint & BRICS retrieves Mobocertinib \\
G & Sulopenem Etzadroxil & Scaffold is sensitive to sulfoxide notation & Functional Group retrieves cephalosporins \\
\bottomrule
\end{tabular}
}
\end{table*}

\begin{figure*}[t]
\centering
\includegraphics[width=\textwidth]{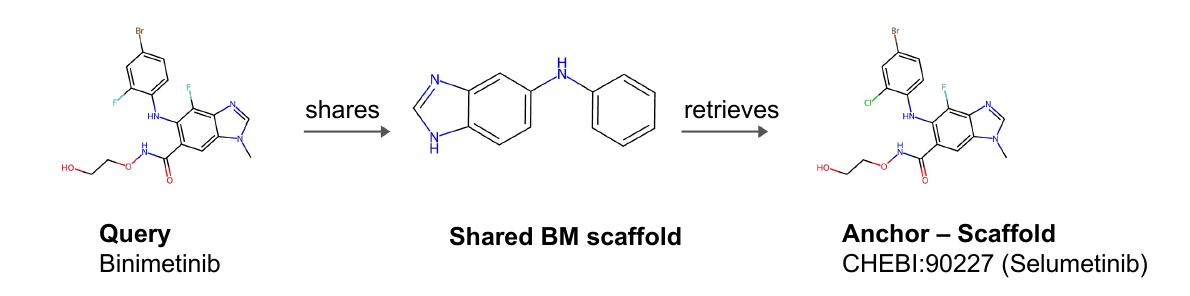}
\caption{\textbf{Case A --- Scaffold identifies a close analogue.} Binimetinib shares its Bemis--Murcko scaffold with \texttt{CHEBI:90227} (Selumetinib), an independently approved MEK1/2 inhibitor with ECFP4 Tanimoto similarity of $0.84$. Selumetinib's own KG relations match four of five gold indications and both gold targets.}
\label{fig:qual_binimetinib}
\end{figure*}

\paragraph{Case A: Scaffold identifies a close analogue (Binimetinib).} Scaffold retrieval identifies \texttt{CHEBI:90227} (Selumetinib), which has the same Bemis--Murcko scaffold as the query (Figure~\ref{fig:qual_binimetinib}). The two molecules are independently approved MEK1/2 inhibitors rather than different records of the same active ingredient; they differ in peripheral halogenation and have ECFP4 Tanimoto similarity of $0.84$. Selumetinib's source relations match four of five gold indications and both gold targets, MAP2K1 and MAP2K2. Fingerprint and BRICS Fragment also identify Selumetinib, whereas Functional Group does not.

\begin{figure*}[t]
\centering
\includegraphics[width=\textwidth]{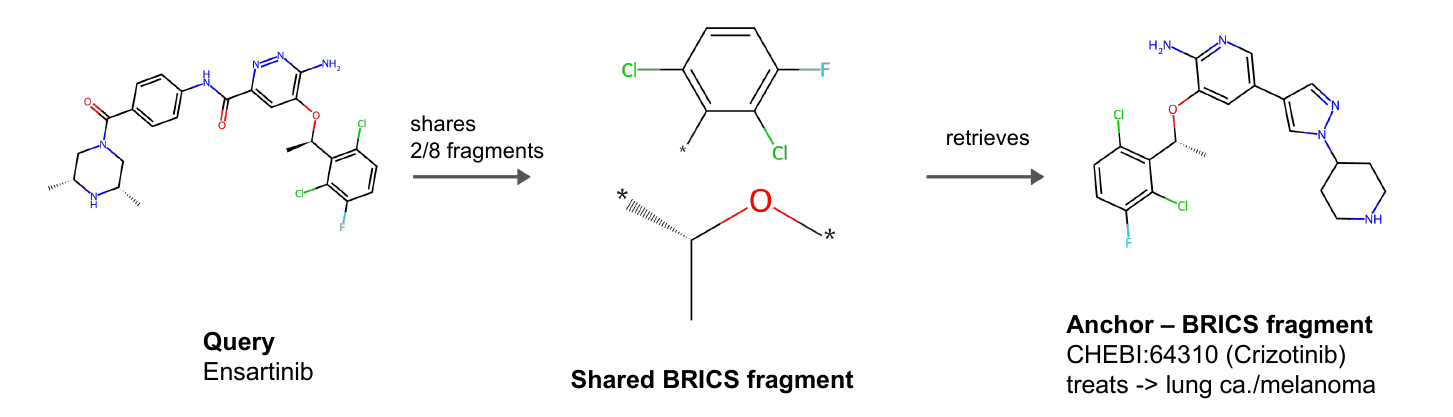}
\caption{\textbf{Case B --- BRICS succeeds without a scaffold match.} Ensartinib has no matching Bemis--Murcko scaffold, but it shares two BRICS fragments with \texttt{CHEBI:64310} (Crizotinib), an independently approved ALK inhibitor with ECFP4 Tanimoto similarity of $0.42$. Crizotinib's own indication and ALK-target relations match gold.}
\label{fig:qual_ensartinib}
\end{figure*}

\paragraph{Case B: BRICS succeeds without a scaffold match (Ensartinib).} Ensartinib has a distinct whole-ring-system scaffold, so whole-scaffold retrieval returns no anchor. BRICS instead isolates two substructures shared with \texttt{CHEBI:64310} (Crizotinib): a 2-chloro-6-fluorophenyl ring and a chiral ether linkage associated with this ALK-inhibitor class (Figure~\ref{fig:qual_ensartinib}). Although their overall ECFP4 Tanimoto similarity is only $0.42$, Crizotinib's source relations match all three gold indications and the gold target ALK. Ensartinib and Crizotinib are independently approved ALK inhibitors, not equivalent records of one compound.

\begin{figure*}[t]
\centering
\includegraphics[width=\textwidth]{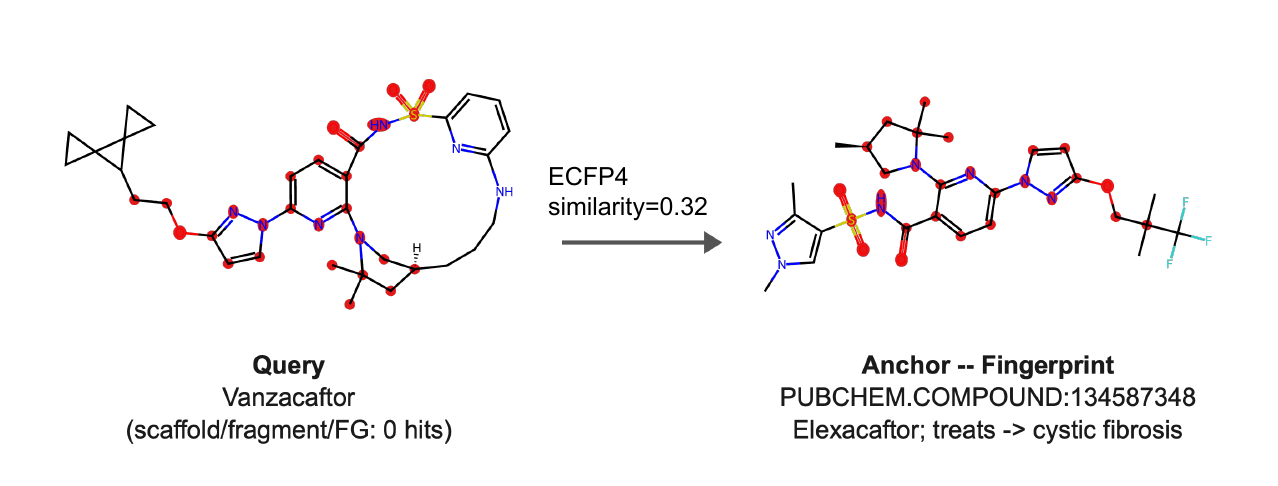}
\caption{\textbf{Case C --- Fingerprint succeeds when local views failed to capture the common structure.} Vanzacaftor has no useful Scaffold, BRICS, or Functional Group anchor shared with any anchor molecules. A diagnostic exhaustive ECFP4 comparison identifies \texttt{PUBCHEM.COMPOUND:134587348} (Elexacaftor) at Tanimoto similarity $0.32$, which shares a huge common structure (highlighted in red) with Vanzacaftor that carries the gold cystic-fibrosis relation.}
\label{fig:qual_vanzacaftor}
\end{figure*}

\paragraph{Case C: Fingerprint succeeds when local views do not (Vanzacaftor).} Scaffold and BRICS return no anchors, while the ten Functional Group candidates carry no cystic-fibrosis indication. A diagnostic, exhaustive ECFP4 comparison across the 2.74 million molecules instead identifies \texttt{PUBCHEM.COMPOUND:134587348} (Elexacaftor), a CFTR corrector in the same combination-therapy class, at Tanimoto similarity $0.32$ (Figure~\ref{fig:qual_vanzacaftor}). Elexacaftor has a different scaffold, fragment set, and functional-group profile, showing that aggregate fingerprint similarity captures a relationship missed by the local views. Its own cystic-fibrosis relation matches gold.

\begin{figure*}[t]
\centering
\includegraphics[width=\textwidth]{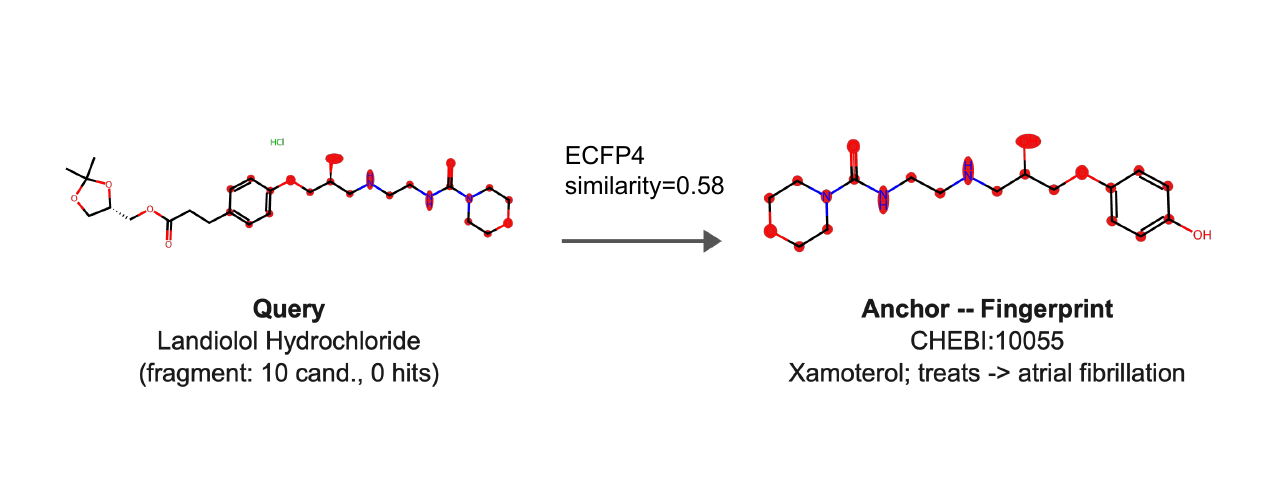}
\caption{\textbf{Case D --- Fingerprint succeeds when fragment matches are uninformative.} Landiolol Hydrochloride's BRICS candidates do not carry the gold indication. ECFP4 similarity identifies \texttt{CHEBI:10055} (Xamoterol), which share a long common structure (highlighted in red), at Tanimoto similarity $0.575$; its atrial-fibrillation relation matches gold, though the two drugs are pharmacologically distinct ($\beta_1$ antagonist vs.\ partial agonist), so the match is disease-level rather than mechanistic.}
\label{fig:qual_landiolol}
\end{figure*}

\paragraph{Case D: Fingerprint succeeds when fragment matches are uninformative (Landiolol Hydrochloride).}
None of the ten BRICS candidates carries the gold atrial-fibrillation indication. Fingerprint retrieval instead identifies \texttt{CHEBI:10055} (Xamoterol), a $\beta_1$-adrenoceptor partial agonist, at Tanimoto similarity $0.575$ (Figure~\ref{fig:qual_landiolol}). This value exceeds the similarity threshold used in the experiments. Landiolol and Xamoterol share an aryloxypropanolamine pharmacophore and a morpholine-containing urea moiety but differ in their peripheral substituents. Their exact Bemis--Murcko scaffolds do not match, and Xamoterol is not recovered among the top BRICS candidates under our retrieval settings. Fingerprint retrieval therefore captures a broader structural similarity missed by BRICS fragment retrieval and recovers the relevant atrial-fibrillation association. However, this disease-level match does not imply mechanistic equivalence: Landiolol is a selective $\beta_1$ antagonist used for rapid rate control, whereas Xamoterol is a $\beta_1$ partial agonist studied for heart-rate stabilization in chronic atrial fibrillation. This case illustrates both the value and the limitation of fingerprint-based evidence transfer.

\begin{figure*}[t]
\centering
\includegraphics[width=\textwidth]{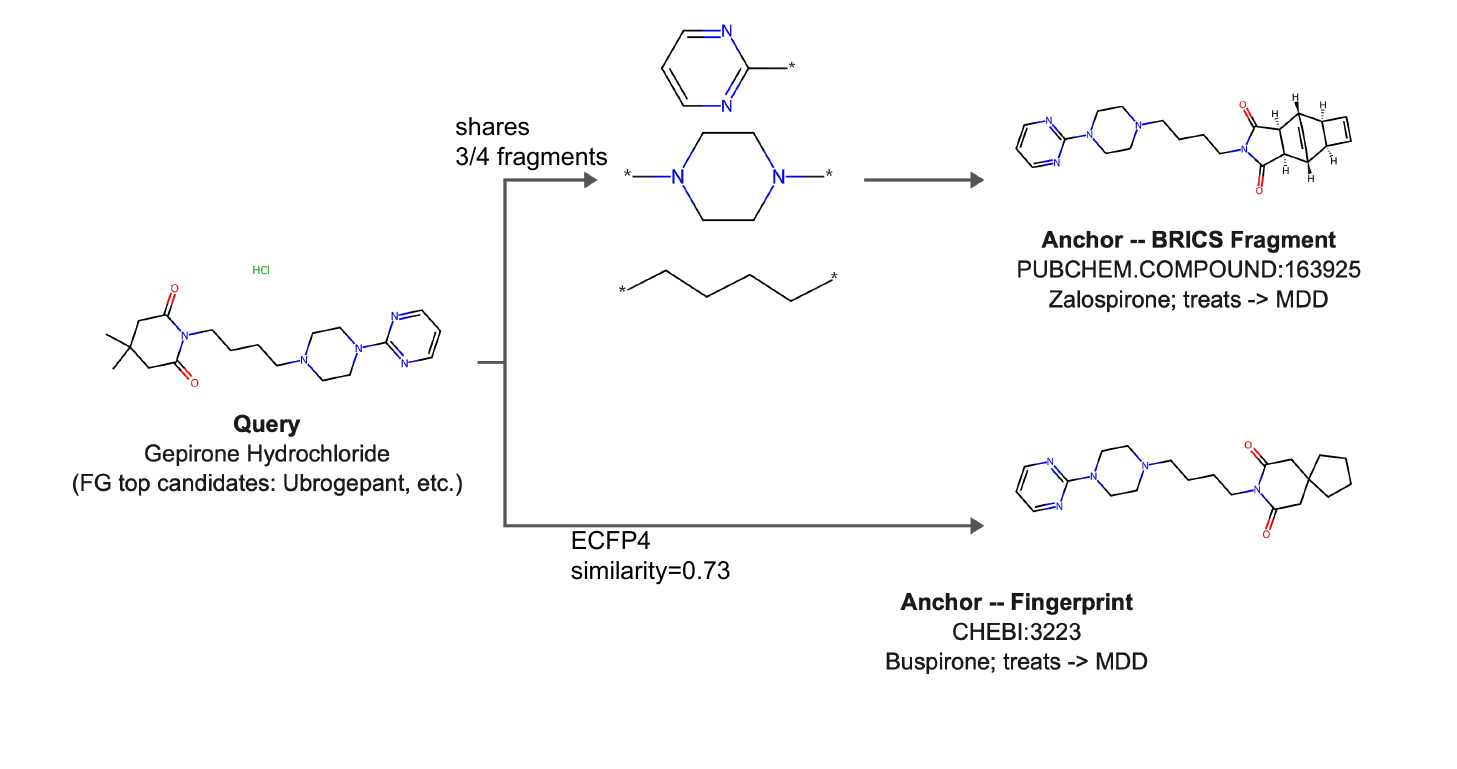}
\caption{\textbf{Case E --- BRICS and Fingerprint overcome nonspecific functional groups.} Gepirone Hydrochloride's highest-ranked Functional Group candidate is pharmacologically unrelated. BRICS identifies Zalospirone through three shared fragments, and Fingerprint identifies Buspirone at Tanimoto similarity $0.73$. Both relevant analogues carry the gold major-depressive-disorder relation.}
\label{fig:qual_gepirone}
\end{figure*}

\paragraph{Case E: BRICS and Fingerprint overcome nonspecific functional groups (Gepirone Hydrochloride).} The highest-ranked Functional Group candidate is \texttt{PUBCHEM.COMPOUND:68748835} (Ubrogepant), a migraine drug that shares generic amide and piperazine groups with the query but is pharmacologically unrelated. Two more specific views independently identify relevant analogues (Figure~\ref{fig:qual_gepirone}). BRICS finds three of four query fragments in \texttt{PUBCHEM.COMPOUND:163925} (Zalospirone), while Fingerprint identifies \texttt{CHEBI:3223} (Buspirone) at Tanimoto similarity $0.73$. Both are members of the azapirone family, and each carries a major-depressive-disorder relation that matches gold.

\begin{figure*}[t]
\centering
\includegraphics[width=\textwidth]{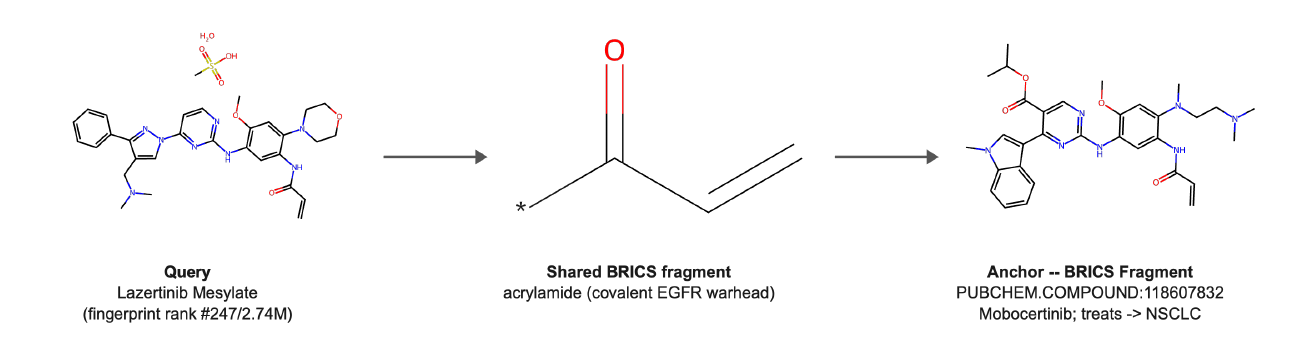}
\caption{\textbf{Case F --- BRICS succeeds when relevant fingerprint matches rank low.} Lazertinib Mesylate's relevant analogues rank 90--247 by ECFP4 similarity. BRICS isolates the shared acrylamide group and identifies \texttt{PUBCHEM.COMPOUND:118607832} (Mobocertinib), whose non-small-cell-lung-carcinoma relation matches gold.}
\label{fig:qual_lazertinib}
\end{figure*}

\paragraph{Case F: BRICS succeeds when relevant fingerprint matches rank low (Lazertinib Mesylate).} Exhaustive ECFP4 comparison places the closest relevant analogues at ranks 90, 100, and 247, with Tanimoto similarities of $0.41$, $0.41$, and $0.37$. These positions fall outside the retrieval cutoff, indicating that whole-molecule similarity dilutes the critical shared feature. BRICS instead isolates the acrylamide group associated with covalent EGFR inhibition (Figure~\ref{fig:qual_lazertinib}) and identifies \texttt{PUBCHEM.COMPOUND:118607832} (Mobocertinib). This analogue carries a non-small-cell-lung-carcinoma relation that matches gold. Here, local decomposition succeeds because the relevant substructure occupies only a small part of two otherwise different molecules.

\begin{figure*}[t]
\centering
\includegraphics[width=\textwidth]{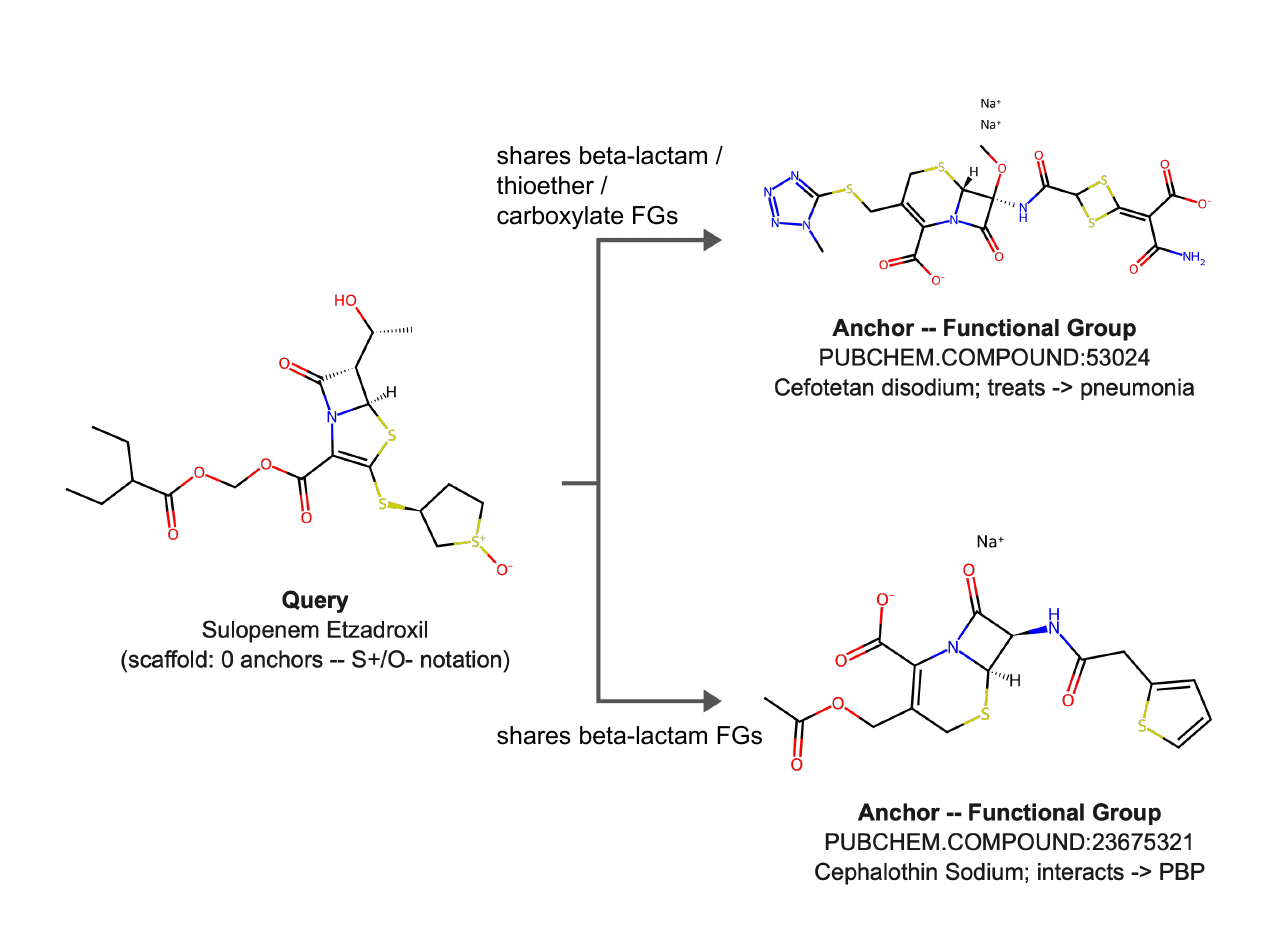}
\caption{\textbf{Case G --- Functional groups remain robust to a notation difference.} Sulopenem Etzadroxil's charge-separated sulfoxide notation prevents Bemis--Murcko scaffold matching against an equivalent neutral representation. Functional Group identifies Cefotetan disodium and Cephalothin Sodium, whose source relations match the gold indication and target.}
\label{fig:qual_sulopenem}
\end{figure*}

\paragraph{Case G: Functional groups remain robust to a notation difference (Sulopenem Etzadroxil).} Scaffold retrieval returns no anchor because the query represents its sulfoxide in charge-separated form (\texttt{S+/O-}), whereas an equivalent molecular record uses neutral \texttt{S(=O)} notation. Bemis--Murcko processing preserves this difference, preventing the scaffolds from matching. Functional Group is less sensitive to the notation and identifies two cephalosporins that share beta-lactam, thioether, and carboxylate groups with the query (Figure~\ref{fig:qual_sulopenem}): \texttt{PUBCHEM.COMPOUND:53024} (Cefotetan disodium) and \texttt{PUBCHEM.COMPOUND:23675321} (Cephalothin Sodium). Their source relations recover the gold pneumonia indication and bacterial penicillin-binding-protein target.

Together, these examples show that no anchor family is uniformly reliable. Scaffold is effective when a conserved core structure is informative, but it can miss a distinct chemotype or become sensitive to molecular notation. BRICS captures small, functionally important substructures that whole-molecule fingerprints may rank too low, whereas Fingerprint can preserve broader similarity when no exact local substructure is shared. Functional Group is robust to some representation differences but can be too nonspecific. This pattern agrees with the anchor-family ablation in the main paper: fusion helps because different queries require different structural views, not because several views estimate the same signal with varying noise.

\end{document}